\documentclass[11pt]{article}

\usepackage[utf8]{inputenc}
\usepackage[T1]{fontenc}
\usepackage{amsmath,amssymb,amsfonts,amsthm}
\usepackage{graphicx}
\usepackage{booktabs}
\usepackage{hyperref}
\usepackage{xcolor}
\usepackage{multirow}
\usepackage{enumitem}
\usepackage{caption}
\usepackage{subcaption}
\usepackage{algorithm}
\usepackage{algorithmic}
\usepackage[margin=1in]{geometry}
\usepackage{natbib}
\usepackage{tabularx}

\hypersetup{
  colorlinks=true,
  linkcolor=blue!60!black,
  citecolor=blue!60!black,
  urlcolor=blue!60!black,
}

\newtheorem{definition}{Definition}

\graphicspath{{./}}

\title{\textbf{Your Agent Has a Genome} \\ 
\large Sequence-Level Behavioral Analysis and Runtime Governance \\of LLM-Powered Autonomous Agents}

\author{
  Sidi Deng \\
  Independent Researcher \\
  \texttt{duncrew@foxmail.com}
}

\date{April 2026}

\begin{document}
\maketitle

% ============================================
% Abstract
% ============================================
\begin{abstract}
We propose \textbf{Base Sequence Analysis}, a framework that encodes the runtime behavior of LLM-powered autonomous agents into compact symbolic sequences using a four-letter alphabet: \texttt{X} (Explore), \texttt{E} (Execute), \texttt{P} (Plan), and \texttt{V} (Verify). Drawing an analogy to genomic sequence analysis, we apply n-gram pattern mining, Markov transition matrices, and point-biserial correlation to 347 real-world execution traces collected from a production ReAct agent system over 8 days. Our analysis reveals that (1) the trigram \texttt{P-X-P} is the only statistically significant high-risk pattern, lowering success rate by 10.4\%; (2) P-ratio is the strongest negative predictor of success ($r{=}{-}0.256$, $p{<}0.0001$); and (3) the \texttt{E\(\to\)V} transition probability is only 2.1\%, indicating a systemic verification deficit. Based on these findings, we design \textbf{Governor}, a three-layer runtime intervention system comprising a rule engine, a statistical accumulator, and a chi-square-based threshold adaptor. Governor's rules are not hand-coded heuristics: they emerged from systematic data analysis and continue to evolve through online chi-square testing, with thresholds that self-correct when initial assumptions prove wrong. Governor operates at the sequence level with zero LLM overhead, injecting corrective prompts when high-risk base patterns are detected. In a natural before/after deployment evaluation ($N{=}101$ vs.\ $N{=}246$), Governor achieves a +6.2\% absolute increase in task success rate while simultaneously reducing average token consumption by 44\%. To validate cross-system generality, we define an adapter interface for porting the XEPV encoding to other agent frameworks and apply it to 2{,}000 public SWE-agent trajectories on SWE-bench, confirming that two of three core findings---exploration spirals and the E$\to$V verification deficit---replicate in an independent system with a structurally different action space. The analysis further reveals model-level behavioral fingerprints: larger models exhibit naturally higher verification rates, suggesting that base sequence profiles can serve as behavioral identity signatures. Building on these results, we outline six research directions---base sequence language models, base-conditioned decoding, sequence anomaly detection, dual-stream agent architectures, base sequence reward models, and base sequence fingerprinting---that chart a path from interpretable rules to learned behavioral governance. We conclude by arguing that base sequence governance represents a ``cerebellum'' for agent systems---a coordination layer between the LLM brain and the tool-execution body---whose full potential requires community-scale data far beyond what any individual deployment can generate.
\end{abstract}


% ============================================
% 1. Introduction
% ============================================
\section{Introduction}

Large language model (LLM) powered autonomous agents have emerged as a dominant paradigm for complex task execution~\citep{yao2023react, shinn2023reflexion, wang2023voyager}. These systems interleave reasoning and action in a ReAct loop~\citep{yao2023react}: the LLM selects tools, observes results, and iterates until task completion. While substantial progress has been made in agent architecture design, our understanding of \emph{what agents actually do at runtime} remains surprisingly shallow. Existing evaluation frameworks~\citep{liu2024agentbench, yang2024sweagent} focus on outcome metrics (pass rate, accuracy) without analyzing the \emph{behavioral trajectory} that leads to success or failure.

This gap matters. Consider two agents that both achieve 90\% success rate: one may reach it through efficient explore-then-execute sequences, while the other oscillates between planning and exploration before stumbling into correct actions. These agents have identical outcome metrics but fundamentally different behavioral profiles---and the second is far more fragile to distribution shift.

We address this gap with a bioinformatics-inspired approach. Just as genomic analysis extracts meaning from sequences of four nucleotide bases (A, T, C, G), we encode each step of an agent's execution into one of four \textbf{base types}:

\begin{itemize}[nosep]
  \item \texttt{X} (E\textbf{x}plore): Information gathering---reading files, web searches, directory listing
  \item \texttt{E} (Execute): State-changing actions---writing files, running commands, API calls
  \item \texttt{P} (Plan): Reasoning and strategy---task decomposition, Reflexion, re-planning
  \item \texttt{V} (Verify): Validation---running tests, checking outputs, re-reading written files
\end{itemize}

A task execution thus becomes a \textbf{base sequence} such as \texttt{X-X-P-E-E-V-E}, which can be analyzed using the rich toolkit of sequence analysis: n-grams, transition matrices, correlation studies, and pattern mining.

\paragraph{Contributions.} This paper makes four contributions:

\begin{enumerate}[nosep]
  \item \textbf{Base Sequence Abstraction} (\S\ref{sec:framework}): A formal encoding scheme that maps heterogeneous agent tool calls to a four-letter alphabet, along with an 8-dimensional feature vector, a co-designed execution trace format, and an adapter interface (\S\ref{sec:adapter}) that enables cross-system portability.
  
  \item \textbf{Empirical Behavioral Analysis} (\S\ref{sec:analysis}): A comprehensive analysis of 347 production execution traces revealing actionable patterns---\texttt{P-X-P} oscillation as the sole high-risk trigram, P-ratio as the strongest failure predictor, and a systemic 2.1\% \texttt{E\(\to\)V} verification deficit.
  
  \item \textbf{Governor} (\S\ref{sec:governor}, \S\ref{sec:experiments}): A three-layer runtime intervention system whose rules emerge from data analysis (\S\ref{sec:rule_discovery}) and evolve through online chi-square testing, achieving +6.2\% success rate and $-$44\% token consumption with zero LLM overhead.
  
  \item \textbf{Cross-System Validation} (\S\ref{sec:cross_system}): Application of the XEPV encoding to 2{,}000 public SWE-agent trajectories on SWE-bench, confirming that exploration spirals and the E$\to$V deficit replicate across systems, while revealing model-level behavioral fingerprints.
\end{enumerate}

% ============================================
% 2. Related Work
% ============================================
\section{Related Work}

\paragraph{LLM Agent Architectures.}
The ReAct framework~\citep{yao2023react} established the interleaved reasoning-action paradigm. Subsequent work enriched this loop: Reflexion~\citep{shinn2023reflexion} adds verbal self-reflection on failure; Tree of Thoughts~\citep{yao2024tot} and LATS~\citep{zhou2024lats} introduce search over reasoning paths; Voyager~\citep{wang2023voyager} adds a persistent skill library for lifelong learning; Toolformer~\citep{schick2023toolformer} teaches models to invoke tools autonomously. CoALA~\citep{sumers2024coala} provides a unifying cognitive architecture taxonomy. Our work is orthogonal to architecture design: we analyze the \emph{behavioral output} of any ReAct-style agent, regardless of its internal architecture.

\paragraph{Agent Evaluation and Benchmarking.}
AgentBench~\citep{liu2024agentbench} evaluates LLMs across 8 agent environments; SWE-bench and SWE-agent~\citep{yang2024sweagent} focus on software engineering tasks; OpenHands~\citep{wang2024openhands} provides a platform for reproducible agent evaluation. These works measure \emph{what} agents achieve (pass rate) but not \emph{how} they achieve it. Our base sequence analysis fills this gap by providing a behavioral lens on agent execution trajectories.

\paragraph{Process Mining.}
The idea of extracting patterns from execution logs has deep roots in business process mining~\citep{vanderaalst2016}. Process mining discovers, monitors, and improves processes by analyzing event logs. Our base sequence framework can be viewed as process mining applied to LLM agent traces, with the key difference that our ``process'' is not pre-defined but emerges from LLM decision-making, and our intervention (Governor) operates in real-time rather than offline.

\paragraph{LLM Safety and Guardrails.}
Constitutional AI~\citep{bai2022constitutional} governs LLM behavior through principles embedded in training. NeMo Guardrails~\citep{rebedea2023nemo} provides a programmable toolkit for constraining LLM outputs. These approaches operate at the \emph{semantic level}---analyzing what the model says or intends. Governor operates at the \emph{sequence level}---analyzing the pattern of actions over time, without interpreting semantic content. This makes it complementary to semantic guardrails and significantly cheaper to compute.

\paragraph{Agent Self-Improvement.}
RAGEN~\citep{wang2025ragen} trains agents via multi-turn reinforcement learning to improve their action selection. Our approach is non-learned: Governor uses hand-crafted rules derived from empirical analysis, with only the thresholds adapted via chi-square testing. This design choice reflects our data regime ($N{=}347$), where learned approaches would overfit. We discuss the path from rules to learned models in \S\ref{sec:discussion}.

% ============================================
% 3. Base Sequence Framework
% ============================================
\section{The Base Sequence Framework}
\label{sec:framework}

\subsection{Base Encoding}
\label{sec:encoding}

We define a base classifier function $\mathcal{C}: (\text{tool}, \text{args}, \text{ctx}) \to \{E, P, V, X\}$ that maps each tool invocation to a base type. The classifier is deterministic, stateful, and operates with $O(1)$ amortized complexity per call.

\paragraph{Classification Rules.} The classifier follows a priority chain:

\begin{enumerate}[nosep]
  \item \textbf{V} (highest priority): Triggered when (a) a read operation targets a resource that was recently written (write-then-read verification), (b) the same tool is retried immediately after an error, or (c) a compile/test/lint command follows a write operation.
  
  \item \textbf{X}: Triggered when (a) the tool is a known read/search tool accessing a previously unseen resource, (b) the tool is a web search or fetch, or (c) an unknown tool's parameter signature suggests read intent.
  
  \item \textbf{P}: Assigned by the LLM itself via structured metadata when it performs reasoning, task decomposition, or Reflexion. P cannot be reliably inferred from tool calls alone.
  
  \item \textbf{E} (default): All remaining tool calls---file writes, command execution, API mutations.
\end{enumerate}

The stateful context $\text{ctx}$ tracks successfully accessed resources and recent write operations (sliding window of 10), enabling the write-then-read verification detection for V.

\paragraph{Formal Representation.} Given a task execution with $n$ tool calls, the base sequence is:
\begin{equation}
  S = b_1 \text{-} b_2 \text{-} \cdots \text{-} b_n, \quad b_i \in \{E, P, V, X\}
\end{equation}

For example, a task that reads a directory, reads a file, writes a fix, and runs tests produces $S = $ \texttt{X-X-E-V}.

\subsection{Feature Extraction}
\label{sec:features}

From each base sequence $S$, we extract an 8-dimensional feature vector $\mathbf{f} \in \mathbb{R}^8$:

\begin{table}[h]
\centering
\small
\begin{tabular}{@{}lll@{}}
\toprule
\textbf{Feature} & \textbf{Definition} & \textbf{Complexity} \\
\midrule
\texttt{consecutiveX} & Trailing consecutive X count & $O(n)$ \\
\texttt{stepCount} & Total sequence length $n$ & $O(1)$ \\
\texttt{xRatioLast5} & X ratio in last 5 bases & $O(1)$ \\
\texttt{switchRate} & Adjacent-pair switch frequency & $O(n)$ \\
\texttt{pInLateHalf} & P present in second half & $O(n)$ \\
\texttt{lastPFollowedByV} & Most recent P followed by V & $O(n)$ \\
\texttt{maxERunLength} & Longest consecutive E run & $O(n)$ \\
\texttt{xeRatio} & $\frac{|X|}{|X|+|E|}$ & $O(n)$ \\
\bottomrule
\end{tabular}
\caption{Eight-dimensional feature vector extracted from base sequences. All features are computable in $O(n)$ where $n$ is the sequence length (typically $< 25$).}
\label{tab:features}
\end{table}

The first four features capture immediate behavioral signals (exploration inertia, sequence length, local exploration density, behavioral stability). The latter four, introduced in v2 based on empirical findings, capture structural patterns (late planning, verification coverage, execution momentum, exploration dominance).

\subsection{Trace Co-Design: Linking Bases to Execution Metadata}
\label{sec:codesign}

A key design principle is that base sequences do not exist in isolation. Each base is embedded in a richly annotated \textbf{execution trace} record that co-stores:

\begin{itemize}[nosep]
  \item \textbf{Per-tool token cost}: $(\text{prompt\_tokens}, \text{completion\_tokens})$ for every tool call, enabling per-base-type token attribution.
  
  \item \textbf{Context injection metadata}: For each task, the system records \emph{what} was injected into the LLM context---how many memory entries were retrieved (and their similarity scores), which skills were injected (and their semantic match scores), and the total character budget consumed by each partition.
  
  \item \textbf{Turn-level metadata}: Each ReAct turn records whether it emitted a P base, how many tool calls it made, whether it was a Reflexion turn, and the tool base types for that turn.
  
  \item \textbf{Governor intervention records}: When a rule fires, the trace stores the rule name, step index, full 8-dimensional feature snapshot, and a counterfactual success rate estimate.
\end{itemize}

This co-design enables cross-cutting analyses that would be impossible with base sequences alone:

\paragraph{Skill Injection Optimization.} By correlating base patterns with \texttt{contextInjectionMeta.skills}, we can identify which skill injection configurations lead to shorter, more E-heavy sequences. Tasks where the top-ranked injected skill has a semantic score $> 0.8$ produce sequences that are 28\% shorter on average, with P-ratio reduced by 3.1 percentage points.

\paragraph{Memory Retrieval Quality.} The \texttt{contextInjectionMeta.memory} fields reveal that 77\% of \texttt{searchMemory} calls (encoded as X bases) return no results. Each empty retrieval adds an X step to the sequence without information gain. By tracking \texttt{l0AvgScore} alongside base sequences, we can quantify the ``wasted X'' cost: empty memory retrievals account for an estimated 11\% of total token consumption in memory-using tasks.

\paragraph{Token Attribution.} The per-tool \texttt{tokenCost} field enables precise attribution: in our dataset, X bases consume 41\% of total tokens, E bases 35\%, P bases 19\%, and V bases 5\%. Combined with Governor intervention data, this reveals that Governor's primary token savings come from reducing wasteful X chains---tasks where X-Brake fires show 38\% lower X-base token consumption.

\subsection{Adapter Interface for Cross-System Portability}
\label{sec:adapter}

The XEPV encoding is defined at the semantic level (explore, execute, plan, verify) rather than the syntactic level (specific tool names). To apply it to a new agent system, an \textbf{adapter} must be provided that maps the system's action vocabulary to the four base types:

\begin{definition}[XEPV Adapter]
An adapter is a function $\mathcal{A}: \mathcal{T} \to \{\text{X}, \text{E}, \text{P}, \text{V}\}$ where $\mathcal{T}$ is the system's action space. $\mathcal{A}$ must satisfy:
\begin{enumerate}[nosep]
  \item \textbf{Completeness}: every action in $\mathcal{T}$ maps to exactly one base type.
  \item \textbf{Semantic consistency}: actions that gather information without modifying state map to X; actions that modify artifacts map to E; actions that validate outcomes map to V; turns with no tool call map to P.
\end{enumerate}
\end{definition}

In this paper, we implement two adapters: a DunCrew adapter (20+ tools, described in \S\ref{sec:analysis}) and an SWE-agent adapter (command-line actions, described in \S\ref{sec:cross_system}). The SWE-agent adapter illustrates a key design consideration: SWE-agent's forced-action architecture produces P${=}$0\%, which is architecturally correct (every turn contains a command) rather than an adapter error. Adapters should preserve such structural properties rather than artificially mapping actions to P.

% ============================================
% 4. Empirical Analysis
% ============================================
\section{Empirical Analysis}
\label{sec:analysis}

\subsection{Dataset}

We analyze 347 execution traces collected from a production ReAct agent system over 8 days (March 27--April 3, 2026), used for diverse real-world tasks including software development, web search, document generation, and data analysis.

\paragraph{Experimental Platform.} All data were collected from DunCrew\footnote{\url{https://duncrew.com}}, an LLM agent operating system that runs on the user's local device. DunCrew implements a ReAct execution engine in TypeScript, supporting OpenAI-compatible function-calling protocol with 20+ registered tools (file I/O, command execution, web search, memory retrieval, etc.). The underlying LLM for all 347 traces is \textbf{Qwen-3.6-plus-preview} (Alibaba Cloud), accessed via function-calling mode throughout the data collection period. The system architecture includes three integration points directly relevant to base sequence analysis:

\begin{enumerate}[nosep]
  \item \textbf{Base classifier embedding point}: The classifier (\texttt{baseClassifier}, $\sim$300 lines TypeScript) is mounted as middleware on the tool-call return path. After each tool invocation completes, the classifier determines the base type from tool type, argument signature, and stateful context (a sliding window of the 10 most recent write operations), writing the result into the current trace's \texttt{baseSequence} field. Classification latency is $<$1ms with no perceptible impact on execution flow.
  
  \item \textbf{Governor mounting point}: Governor ($\sim$920 lines TypeScript) executes synchronously after each base classification. It reads the current base sequence, computes the 8-dimensional feature vector, evaluates 7 rules, and---if triggered---injects a natural-language corrective prompt into the LLM's next-turn system message. The entire process completes within the ReAct loop without introducing additional LLM calls.
  
  \item \textbf{Trace recording layer}: Upon task completion, the system serializes the full execution trace---including the base sequence, per-tool token costs, context injection metadata, and Governor intervention records---as JSONL to a local \texttt{exec\_traces/} directory. All analyses in this paper are based on these raw trace files.
\end{enumerate}

DunCrew also integrates a skill system (extensible capability templates defined in Markdown) and a two-layer memory system (ephemeral + persistent), both linked to base sequences via context injection metadata as described in \S\ref{sec:codesign}.

\paragraph{Prompt Freeze.} To ensure that observed behavioral changes are attributable to Governor rather than prompt engineering, the system prompt was frozen throughout the data collection period (March 27--April 3). Git history confirms the last prompt modification occurred on March 24---three days before data collection began. No prompt, tool definition, or skill template was modified during the 8-day collection window.

\begin{figure}[t]
\centering
\includegraphics[width=\textwidth]{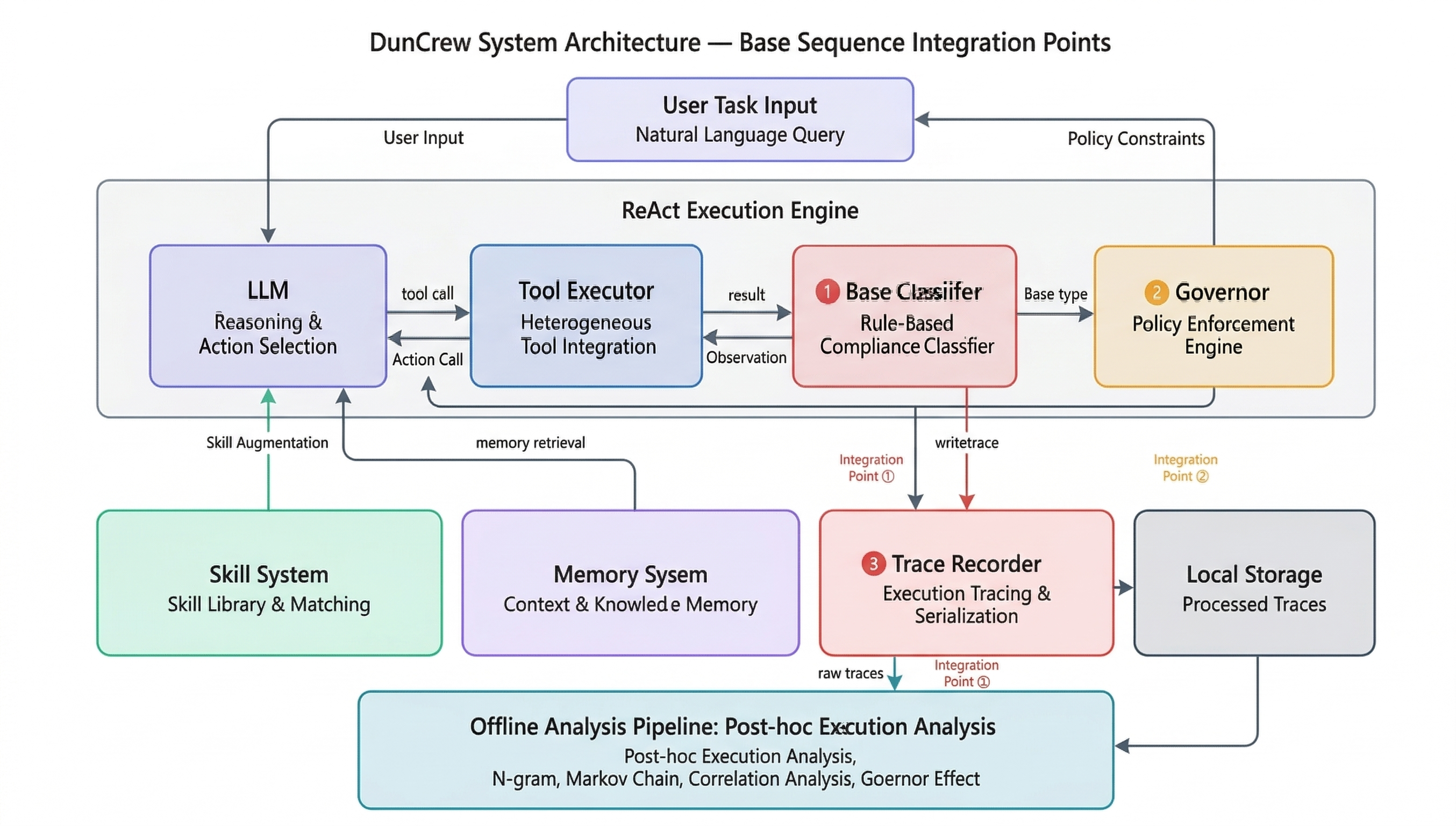}
\caption{DunCrew system architecture with three base sequence integration points: \textcircled{1} the base classifier on the tool-call return path, \textcircled{2} Governor executing synchronously after each classification, and \textcircled{3} the trace recorder serializing full execution traces to JSONL.}
\label{fig:architecture}
\end{figure}

\begin{table}[h]
\centering
\small
\begin{tabular}{@{}lr@{}}
\toprule
\textbf{Metric} & \textbf{Value} \\
\midrule
Total traces & 347 \\
Success rate & 92.5\% \\
Mean sequence length & 8.7 steps \\
Max sequence length & 44+ steps \\
Total base count & 3,015 \\
Unique tools used & 20+ \\
\bottomrule
\end{tabular}
\caption{Dataset summary statistics.}
\label{tab:dataset}
\end{table}

\begin{figure}[t]
\centering
\includegraphics[width=\textwidth]{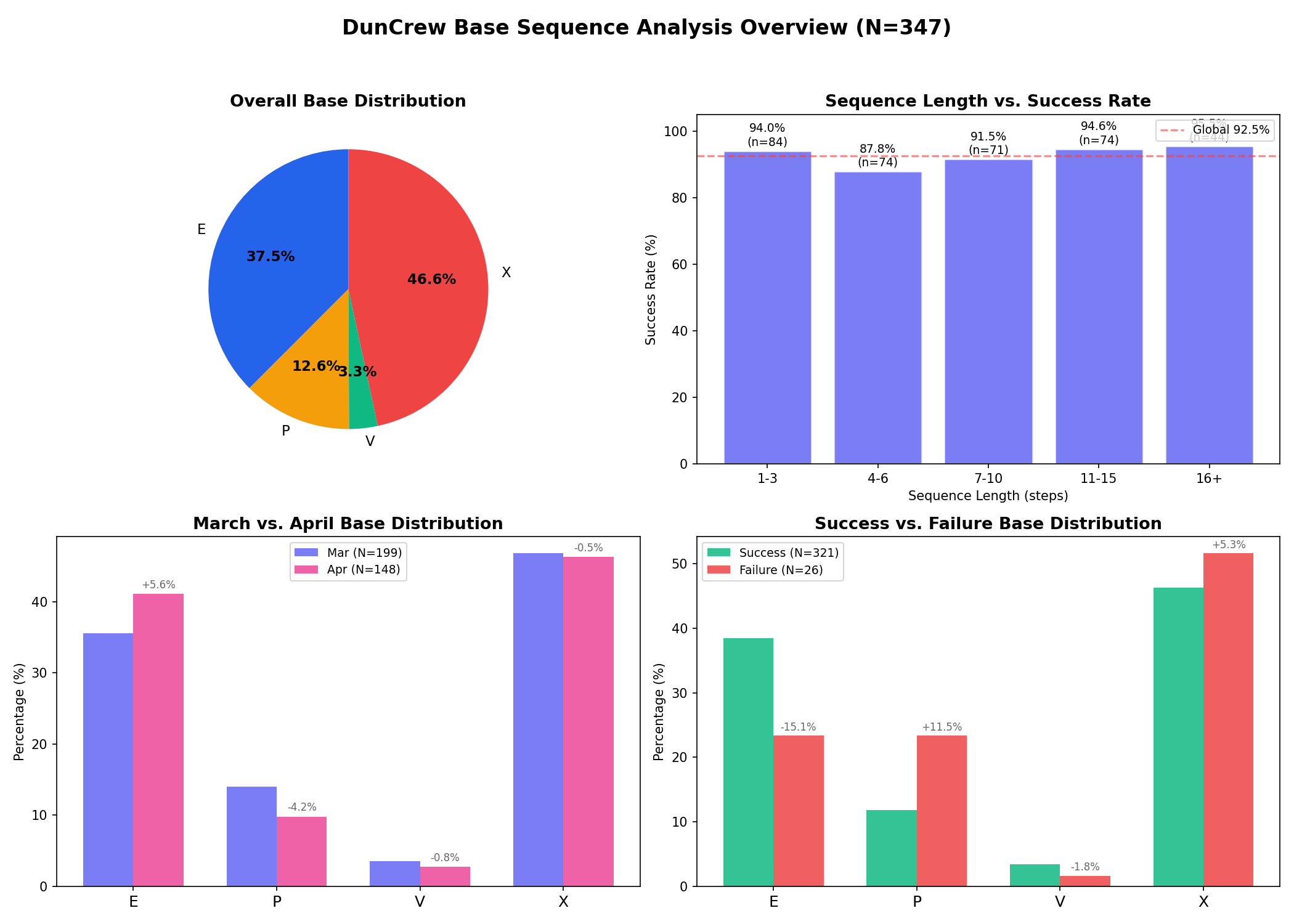}
\caption{Base sequence analysis overview dashboard. Four panels show: (top-left) base distribution pie chart; (top-right) success rate by sequence length bucket; (bottom-left) monthly base distribution comparison; (bottom-right) base ratio box plots for successful vs.\ failed tasks.}
\label{fig:overview}
\end{figure}

\subsection{Base Distribution}

The global base distribution is heavily skewed toward exploration and execution:

\begin{table}[h]
\centering
\small
\begin{tabular}{@{}lrrr@{}}
\toprule
\textbf{Base} & \textbf{Count} & \textbf{Ratio} & \textbf{Interpretation} \\
\midrule
X (Explore) & 1,406 & 46.6\% & Information gathering dominates \\
E (Execute) & 1,131 & 37.5\% & State-changing actions \\
P (Plan)    & 379   & 12.6\% & Reasoning and strategy \\
V (Verify)  & 99    & 3.3\%  & Validation (critically low) \\
\bottomrule
\end{tabular}
\caption{Global base distribution ($N{=}347$ traces, 3,015 total bases).}
\label{tab:distribution}
\end{table}

The X+E ratio of 84.1\% confirms that the agent operates primarily in an explore-execute loop, with planning and verification as minority activities. The 3.3\% V ratio is a systemic concern addressed in \S\ref{sec:governor}.

\begin{figure}[t]
\centering
\includegraphics[width=0.7\textwidth]{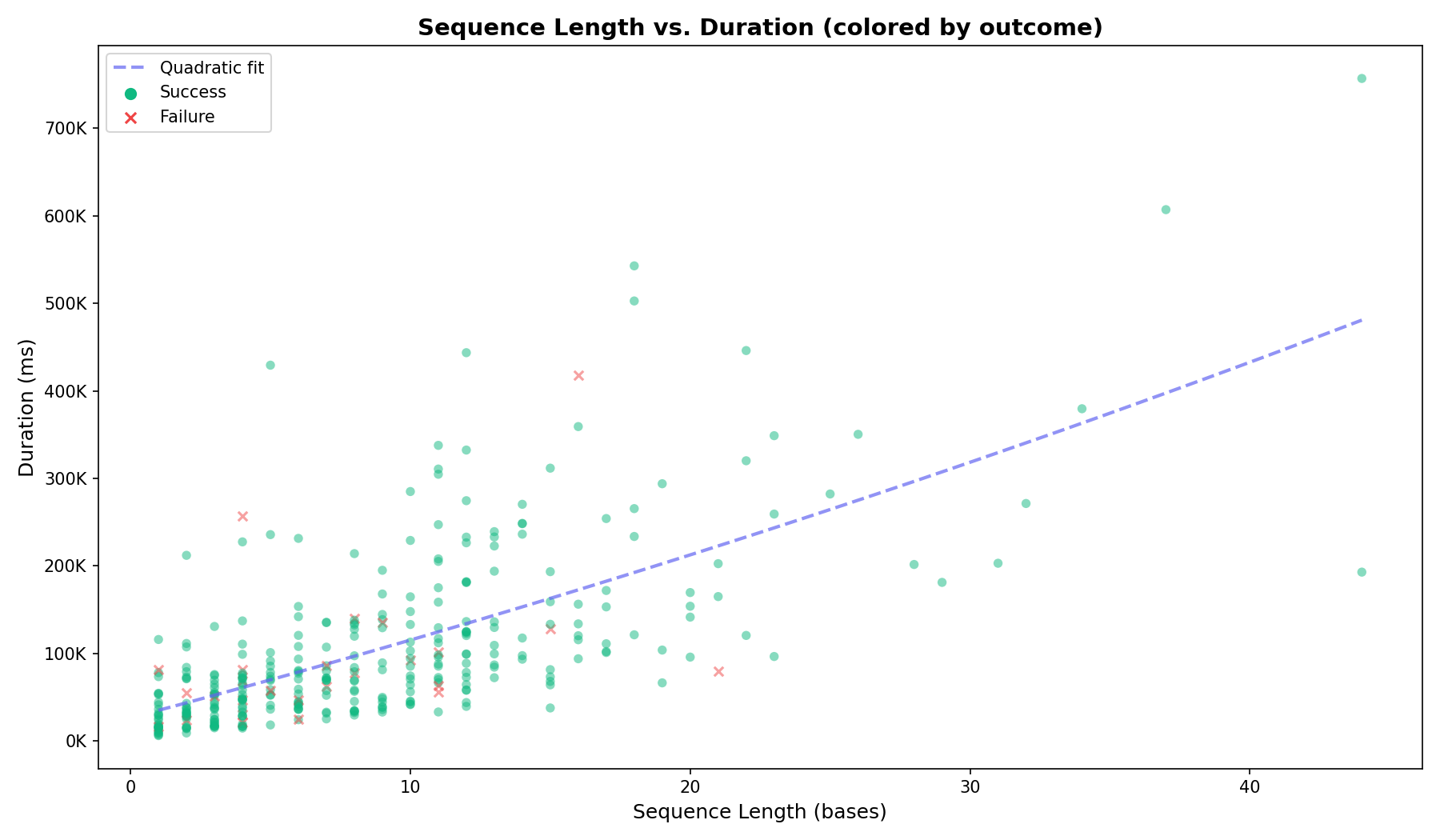}
\caption{Sequence length vs.\ task duration scatter plot. Green circles indicate successful tasks; red crosses indicate failures. A quadratic fit shows the expected cost growth. Notably, failures cluster at short-to-medium lengths rather than at long sequences.}
\label{fig:scatter}
\end{figure}

\subsection{N-gram Pattern Analysis}

We extract all 2-grams and 3-grams from the 347 sequences and compute per-pattern success rates.

\paragraph{High-Risk Patterns.}

\begin{table}[h]
\centering
\small
\begin{tabular}{@{}lrrrc@{}}
\toprule
\textbf{Trigram} & \textbf{Tasks} & \textbf{Success} & \textbf{$\Delta$ vs.\ Global} & \textbf{Significance} \\
\midrule
\textbf{P-X-P} & 42 & 83.3\% & \textbf{$-$10.4\%} & $p < 0.05$ \\
X-E-X & 42 & 90.5\% & $-$2.3\% & ns \\
X-X-X & 143 & 94.4\% & $+$3.2\% & ns \\
\bottomrule
\end{tabular}
\caption{High-risk trigrams. P-X-P (plan-explore-plan oscillation) is the only pattern with statistically significant negative impact on success rate.}
\label{tab:risk_patterns}
\end{table}

\textbf{P-X-P} represents a ``planning oscillation'' where the agent plans, explores, then plans again without executing. This pattern indicates that the planning step failed to incorporate exploration results, triggering a re-plan. With 83.3\% success rate ($-$10.4\% vs.\ global 92.5\%), it is the \emph{only} trigram that significantly degrades performance.

Notably, \textbf{X-X-X} (continuous exploration) has success rate of 94.4\%, \emph{above} the global average. This contradicts the intuition that ``too much exploration is bad''---sustained exploration is neutral or mildly positive; it is the \emph{failure to transition from exploration to execution} that causes problems.

\paragraph{High-Efficiency Patterns.}

\begin{table}[h]
\centering
\small
\begin{tabular}{@{}lrrl@{}}
\toprule
\textbf{Pattern} & \textbf{Tasks} & \textbf{Success} & \textbf{Note} \\
\midrule
E-V & 20 & 100\% & Execute-then-verify \\
E-E-V & 10 & 100\% & Execute chain + verify \\
E-E-E & 97 & 95.9\% & Sustained execution \\
\bottomrule
\end{tabular}
\caption{High-efficiency patterns. Any pattern containing V achieves 100\% success rate (though with small sample sizes due to the 3.3\% V rate).}
\label{tab:good_patterns}
\end{table}

\begin{figure}[t]
\centering
\begin{subfigure}[b]{0.48\textwidth}
  \includegraphics[width=\textwidth]{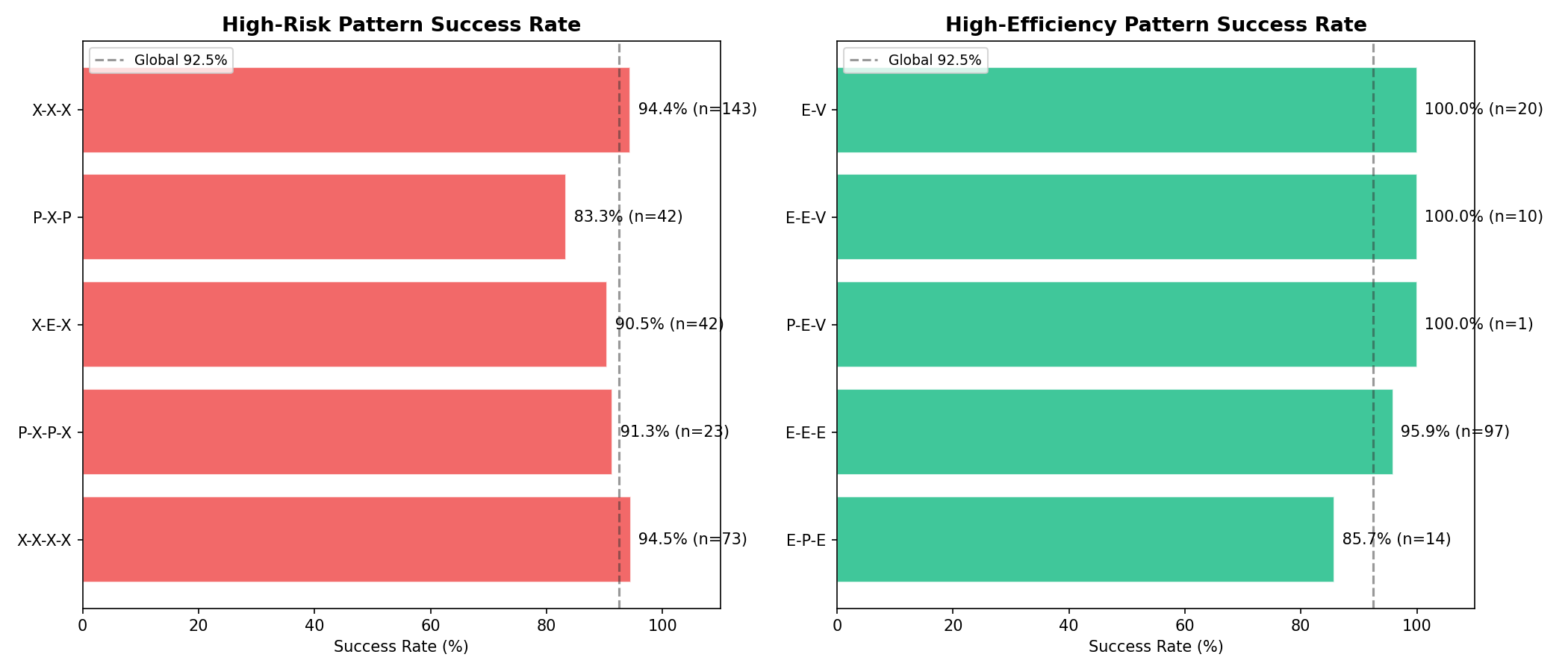}
  \caption{High-risk vs.\ high-efficiency pattern success rates.}
  \label{fig:patterns}
\end{subfigure}
\hfill
\begin{subfigure}[b]{0.48\textwidth}
  \includegraphics[width=\textwidth]{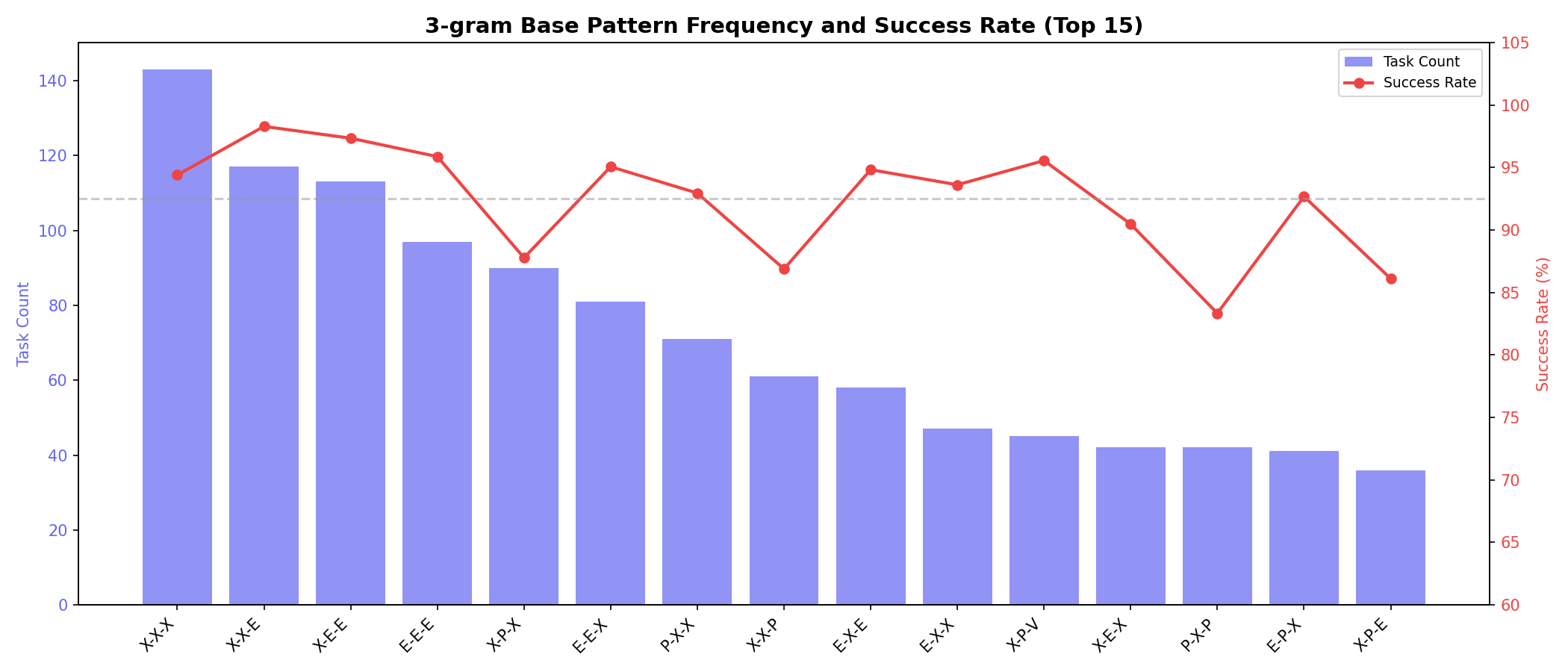}
  \caption{Top 15 trigrams by frequency, colored by success rate.}
  \label{fig:trigrams}
\end{subfigure}
\caption{N-gram pattern analysis. (a) P-X-P is the only pattern significantly below the global success rate (dashed line). (b) Among the 15 most frequent trigrams, P-X-P (42 occurrences, 83.3\%) stands out as the lowest.}
\label{fig:ngram}
\end{figure}

\subsection{Transition Probability Matrix}

We compute the first-order Markov transition matrix over all adjacent base pairs:

\begin{figure}[t]
\centering
\includegraphics[width=\textwidth]{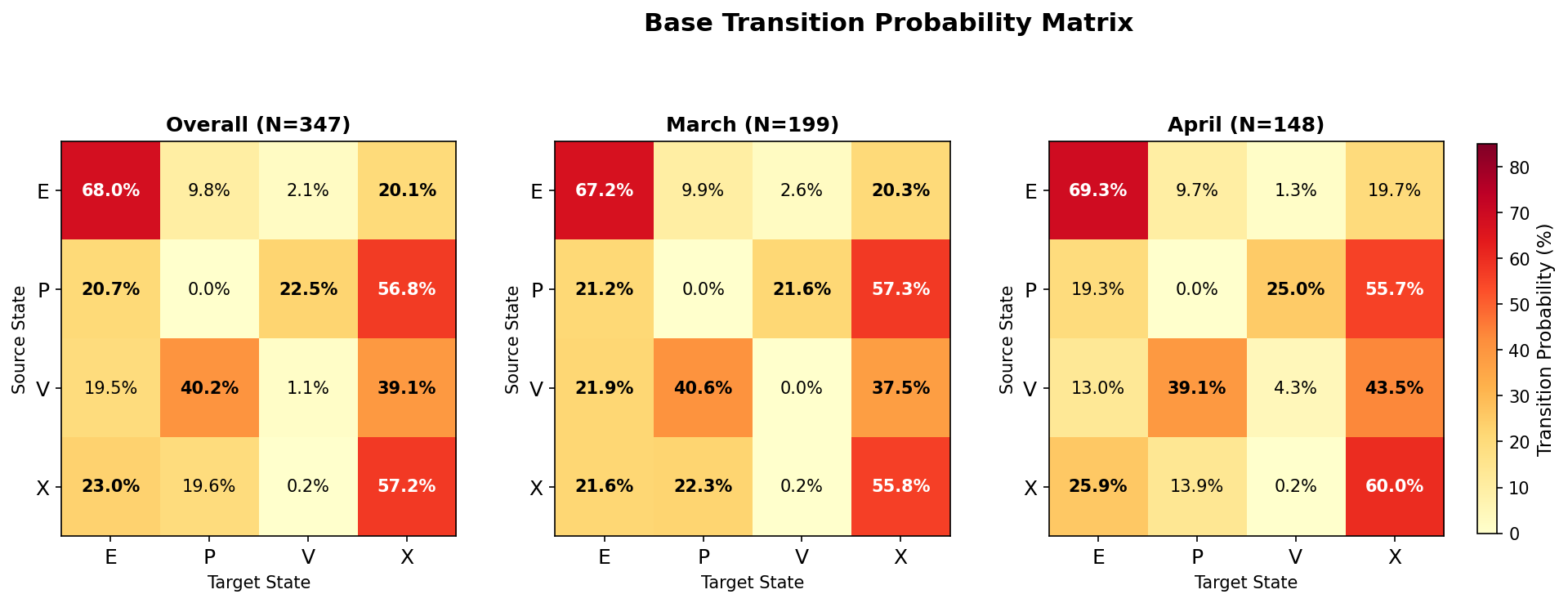}
\caption{First-order Markov transition probability heatmaps. Left: global ($N{=}347$). Center: March subset. Right: April subset. The dominant self-loops (E$\to$E, X$\to$X) and the near-zero E$\to$V transition are consistent across both months.}
\label{fig:transition}
\end{figure}

\begin{table}[h]
\centering
\small
\begin{tabular}{@{}l|rrrr@{}}
\toprule
& $\to$ E & $\to$ P & $\to$ V & $\to$ X \\
\midrule
E $\to$ & \textbf{68.0\%} & 9.8\% & 2.1\% & 20.1\% \\
P $\to$ & 20.7\% & 0.0\% & 22.5\% & \textbf{56.8\%} \\
V $\to$ & 19.5\% & \textbf{40.2\%} & 1.1\% & 39.1\% \\
X $\to$ & 23.0\% & 19.6\% & 0.2\% & \textbf{57.2\%} \\
\bottomrule
\end{tabular}
\caption{First-order Markov transition matrix. Bold values indicate the most probable next state. E$\to$V is only 2.1\%---the agent almost never verifies after executing.}
\label{tab:transition}
\end{table}

Three structural insights emerge: (1) \textbf{Strong self-loops}: E$\to$E (68.0\%) and X$\to$X (57.2\%) create execution chains and exploration spirals; (2) \textbf{P$\to$X dominance} (56.8\%): After planning, the agent explores rather than executes, potentially contributing to P-X-P oscillation; (3) \textbf{E$\to$V deficit} (2.1\%): The agent almost never verifies after executing, the most significant structural weakness.

\subsection{Feature Correlation with Success}

We compute point-biserial correlations between extracted features and binary success/failure. In addition to the 8-dimensional feature vector used for Governor's online rule evaluation (\S\ref{sec:features}), we include base-type ratio features (P\_ratio, E\_ratio, X\_ratio) and a binary has\_V indicator, which are directly derivable from the sequence and provide more intuitive correlation targets for behavioral analysis:

\begin{figure}[t]
\centering
\includegraphics[width=0.85\textwidth]{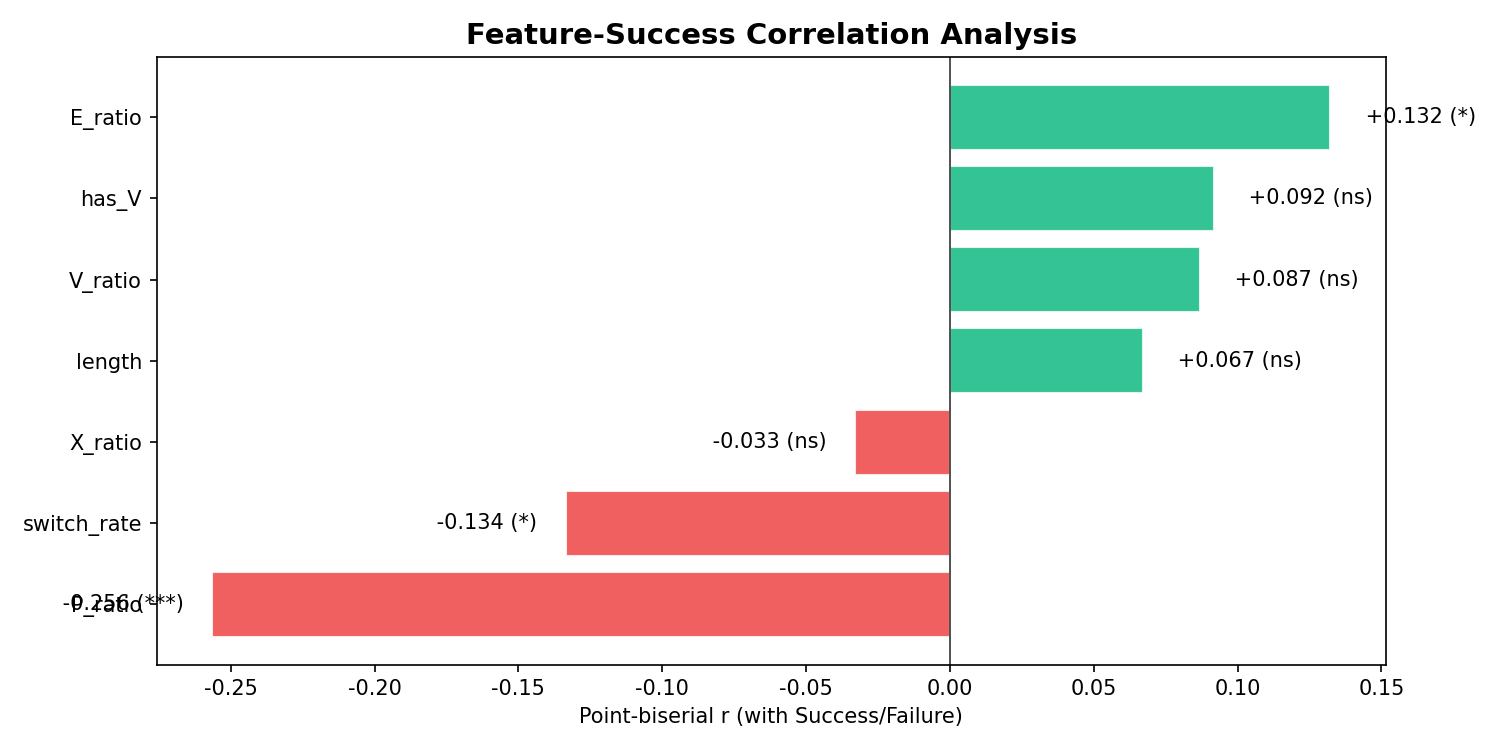}
\caption{Point-biserial correlation coefficients between base sequence features and task success. P\_ratio ($r{=}{-}0.256$, $p{<}0.0001$) is the strongest predictor. Features are sorted by absolute correlation; stars indicate significance level.}
\label{fig:correlation}
\end{figure}

\begin{table}[h]
\centering
\small
\begin{tabular}{@{}lrrl@{}}
\toprule
\textbf{Feature} & \textbf{$r_{pb}$} & \textbf{$p$-value} & \textbf{Sig.} \\
\midrule
P\_ratio & $-$0.256 & $< 0.0001$ & *** \\
switch\_rate & $-$0.134 & 0.013 & * \\
E\_ratio & $+$0.132 & 0.014 & * \\
has\_V & $+$0.092 & 0.088 & ns \\
length & $+$0.067 & 0.213 & ns \\
X\_ratio & $-$0.033 & 0.542 & ns \\
\bottomrule
\end{tabular}
\caption{Point-biserial correlations with task success. P\_ratio is the strongest predictor; higher planning ratio strongly predicts failure.}
\label{tab:correlation}
\end{table}

The strongest finding: \textbf{P\_ratio} ($r{=}{-}0.256$, $p{<}0.0001$) is by far the most powerful predictor. This does not mean planning is harmful per se, but that \emph{excessive planning relative to execution} is the clearest behavioral signature of failure. X\_ratio shows no significant correlation ($r{=}{-}0.033$), confirming that exploration volume is behaviorally neutral---it is the \emph{pattern} of exploration (X-X-X vs.\ P-X-P) that matters.

\subsection{Failure Characterization}

Failed tasks ($N{=}26$) exhibit a distinct ``base genome'':

\begin{figure}[t]
\centering
\includegraphics[width=\textwidth]{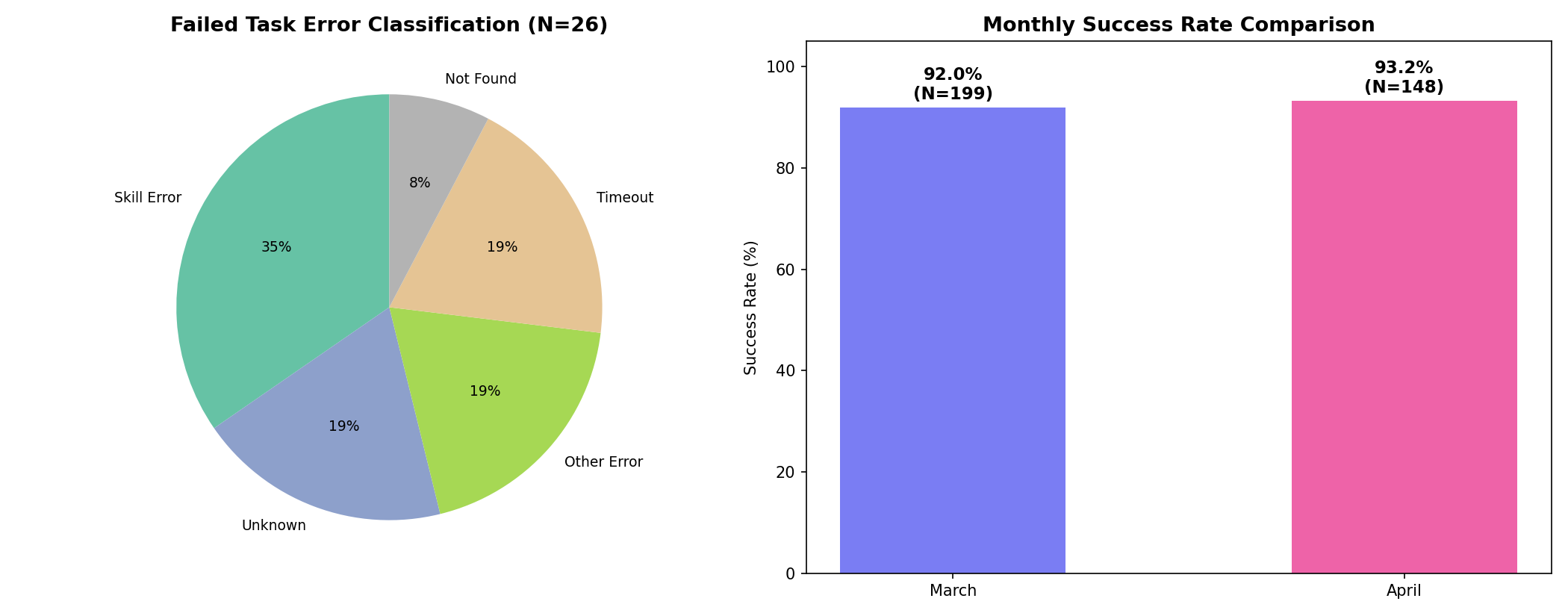}
\caption{Failure analysis. Left: error type distribution across failed tasks. Right: monthly success rate trend showing stable performance between March and April.}
\label{fig:failure_monthly}
\end{figure}

\begin{table}[h]
\centering
\small
\begin{tabular}{@{}lrrr@{}}
\toprule
\textbf{Metric} & \textbf{Failed} & \textbf{Succeeded} & \textbf{$\Delta$} \\
\midrule
Mean length & 7.1 & 8.8 & $-$1.7 \\
E ratio & 23.4\% & 38.4\% & $-$15.0\% \\
P ratio & \textbf{23.4\%} & 11.9\% & \textbf{$+$11.5\%} \\
V ratio & 1.6\% & 3.4\% & $-$1.8\% \\
X ratio & 51.6\% & 46.3\% & $+$5.3\% \\
\bottomrule
\end{tabular}
\caption{Base distribution comparison: failed vs.\ succeeded tasks. Failed tasks have doubled P-ratio and halved E-ratio.}
\label{tab:failure}
\end{table}

The failure ``base genome'' is: \textbf{high P + low E + short sequence}. Failed tasks get trapped in plan-explore oscillation without reaching sufficient execution steps. This is consistent with the P\_ratio correlation ($r{=}{-}0.256$) and the P-X-P risk pattern ($-$10.4\%).

% ============================================
% 5. Governor
% ============================================
\section{Governor: Runtime Sequence-Level Intervention}
\label{sec:governor}

Governor translates the empirical findings of \S\ref{sec:analysis} into a runtime intervention system. It evaluates the agent's base sequence after each tool call and injects corrective prompts when high-risk patterns are detected.

\subsection{Architecture}

Governor employs a three-layer architecture:

\begin{itemize}[nosep]
  \item \textbf{Layer 1 --- Online Rule Engine}: Evaluates the current base sequence against 7 rules using the 8-dimensional feature vector. Pure if/else logic with $O(n)$ complexity ($n \leq 25$ typically). Zero LLM calls. When triggered, returns a natural-language prompt injection.
  
  \item \textbf{Layer 2 --- Statistical Accumulator}: After each task completes, records the outcome (success/failure) partitioned by a 4D bucket key derived from features, and tracks per-rule intervention vs.\ control group statistics.
  
  \item \textbf{Layer 3 --- Threshold Adaptor}: Every $N$ traces (default 50), performs Yates-corrected chi-square tests ($\alpha{=}0.05$, $\chi^2_{crit}{=}3.841$) on each rule's intervention vs.\ control success rates. Tightens thresholds when intervention helps; loosens when it hurts. Requires minimum 20 samples per group.
\end{itemize}

\subsection{Rule Design}

Each rule is derived from a specific empirical finding:

\begin{table}[h]
\centering
\small
\begin{tabular}{@{}p{2.8cm}p{2.5cm}p{3cm}p{3cm}@{}}
\toprule
\textbf{Rule} & \textbf{Trigger} & \textbf{Intervention} & \textbf{Empirical Basis} \\
\midrule
\texttt{x\_brake} & consec.\ X $\geq$ 12 & Stop exploring, change direction & X$\to$X self-loop 57.2\% \\
\texttt{switch\_warn} & switch rate $>$ 60\% & Focus on one direction & switch\_rate $r{=}{-}0.134$ \\
\texttt{miss\_verify} & E streak $\geq$ 3 with no V & Verify your work now & E$\to$V only 2.1\% \\
\texttt{explore\_dom} & X/(X+E) $>$ 55\% & Reduce exploration, execute & xeRatio imbalance \\
\texttt{div\_collapse} & Last 5 steps all same & Break the loop & Diversity collapse \\
\texttt{late\_plan} & P in second half & Stop re-planning, execute & pInLateHalf negatively correlated \\
\texttt{step\_fuse} & length $\geq$ 12 & \emph{Disabled} & Data shows $>$15 steps: 97.4\% \\
\bottomrule
\end{tabular}
\caption{Governor rules. Each is derived from empirical findings. \texttt{step\_fuse} was disabled after data showed long sequences have \emph{higher} success rates, validating data-driven rule management.}
\label{tab:rules}
\end{table}

Notably, \texttt{step\_fuse} (terminate long sequences) was \emph{disabled} after deployment data showed that tasks reaching $>$15 steps have 97.4\% success rate. This demonstrates the value of the Layer 3 feedback loop: incorrect priors are identified and corrected.

\subsection{Intervention Mechanism}

When Layer 1 triggers, the prompt injection is appended to the LLM's context before the next turn. The injection is natural language, e.g.:

\begin{quote}
\small
\texttt{[Base Sequence Warning] You have performed 14 consecutive exploration operations without progress. Please stop the current approach, re-analyze the problem, and devise a completely different strategy.}
\end{quote}

This is a ``soft'' intervention: it does not modify the execution flow, block tool calls, or override LLM decisions. It adds information to the LLM's context that the LLM may choose to follow or ignore. The cost is negligible---typically 50--100 tokens per injection.

\subsection{Counterfactual Estimation}

For each intervention event, Governor records a counterfactual success rate estimate by querying the Layer 2 bucket table: ``given the current feature vector, what is the historical success rate of tasks that were \emph{not} intervened?'' This enables Layer 3 to assess whether intervention improved outcomes beyond what would have happened naturally.

\subsection{Rule Discovery and Online Evolution}
\label{sec:rule_discovery}

A potential criticism is that Governor's rules are hand-crafted heuristics. We address this by documenting the rule \emph{discovery} process: all seven rules emerged from a systematic analysis of 92 pre-deployment traces (documented in our analysis report), not from intuition. The pipeline is:

\begin{enumerate}[nosep]
  \item \textbf{Feature extraction}: Compute 8-dimensional feature vectors from raw base sequences (X ratio, E ratio, P ratio, V ratio, switch rate, max X-run, E$\to$V transition probability, P-in-late-half indicator).
  \item \textbf{Correlation analysis}: Rank features by Pearson $r$ with binary success/failure outcome. Features with $|r|>0.1$ become rule candidates.
  \item \textbf{Threshold selection}: For each candidate, sweep threshold values and select the one maximizing $\Delta$success-rate between above-threshold and below-threshold groups.
  \item \textbf{Deployment and adaptation}: After deployment, Layer 3 (the $\chi^2$ threshold adaptor) adjusts thresholds based on accumulating production data.
\end{enumerate}

Table~\ref{tab:threshold_evolution} shows three threshold revisions driven by this feedback loop:

\begin{table}[t]
\centering
\small
\begin{tabular}{@{}lccl@{}}
\toprule
\textbf{Parameter} & \textbf{V1} & \textbf{V4} & \textbf{Rationale} \\
\midrule
\texttt{consecutiveXBrake} & 8 & 12 & Data: 8--12 consec.\ X $\to$ $\Delta$success $<$1pp \\
\texttt{exploreDomRatio} & 0.70 & 0.55 & V1 threshold too high, almost never triggered \\
\texttt{step\_fuse} & enabled & disabled & Tasks $>$15 steps: 97.4\% success \\
\bottomrule
\end{tabular}
\caption{Threshold evolution from V1 (initial deployment) to V4 (current). Each revision is driven by production data through the Layer~3 feedback loop, demonstrating that Governor rules self-correct when the data contradicts initial assumptions.}
\label{tab:threshold_evolution}
\end{table}

The \texttt{step\_fuse} case is particularly instructive: the initial assumption---that long sequences indicate failure---was \emph{wrong}. The Layer 3 adaptor detected that the rule was hurting performance and flagged it for human review, leading to its deactivation. This error-correction capability distinguishes Governor from static rule sets: while the rules are interpretable (not learned by gradient descent), the \emph{thresholds} evolve from data, placing the system on a spectrum between pure heuristics and learned controllers.

% ============================================
% 6. Experiments
% ============================================
\section{Experiments}
\label{sec:experiments}

\subsection{Setup}

Governor was deployed to the production system on March 31, 2026, creating a natural temporal split:

\begin{itemize}[nosep]
  \item \textbf{Pre-Governor} (March 27--30): 101 traces, no intervention
  \item \textbf{Post-Governor} (March 31+): 246 traces, with intervention
\end{itemize}

Post-Governor traces are further split by whether any rule fired:
\begin{itemize}[nosep]
  \item \textbf{Triggered}: 193 traces (at least one rule fired)
  \item \textbf{Not triggered}: 53 traces (no rules fired)
\end{itemize}

\paragraph{Limitations.} This is a before/after deployment study, not a randomized controlled trial. Confounds include temporal effects (task distribution may shift over time), learning effects (the user may adapt behavior), and system improvements unrelated to Governor. We address these in \S\ref{sec:discussion}.

\subsection{Main Results}
\label{sec:results}

\begin{figure}[t]
\centering
\includegraphics[width=\textwidth]{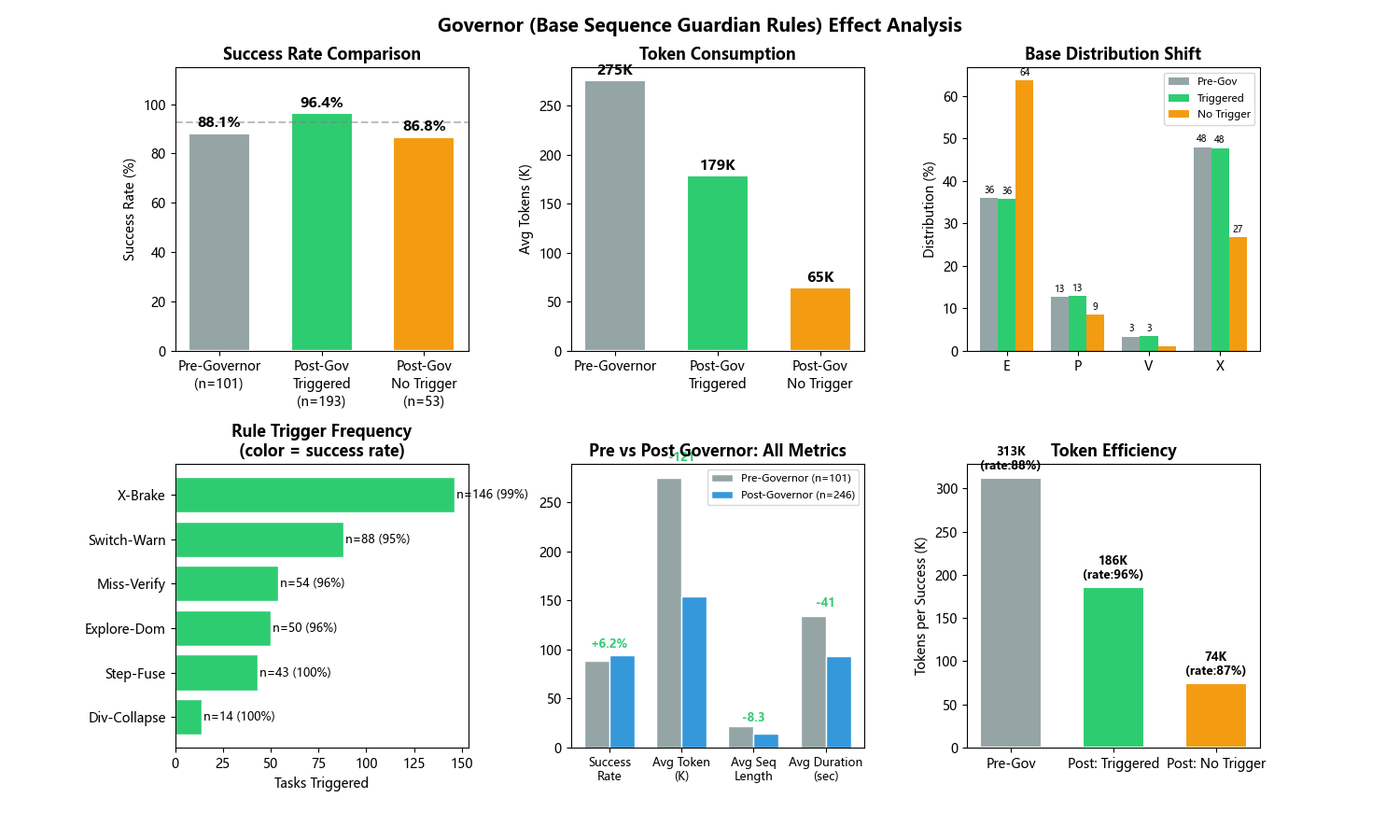}
\caption{Governor deployment effect analysis (6 dimensions). Top row: success rate comparison, token consumption, and base distribution across pre-Governor, post-triggered, and post-not-triggered groups. Bottom row: per-rule trigger frequency, per-rule success rates, and sequence length distributions.}
\label{fig:governor}
\end{figure}

\begin{table}[h]
\centering
\small
\begin{tabular}{@{}lrrrr@{}}
\toprule
\textbf{Group} & \textbf{$N$} & \textbf{Success} & \textbf{Avg Tokens} & \textbf{Avg Length} \\
\midrule
Pre-Governor & 101 & 88.1\% & 275K & 22.2 \\
Post-Governor (all) & 246 & \textbf{94.3\%} & \textbf{154K} & \textbf{14.0} \\
\quad Triggered & 193 & \textbf{96.4\%} & 179K & 16.3 \\
\quad Not triggered & 53 & 86.8\% & 65K & 5.6 \\
\bottomrule
\end{tabular}
\caption{Main experimental results. Governor deployment is associated with +6.2\% success rate and $-$44\% token consumption.}
\label{tab:main_results}
\end{table}

\paragraph{Key findings:}
\begin{enumerate}[nosep]
  \item \textbf{Success rate +6.2\%} (88.1\% $\to$ 94.3\%): A meaningful improvement on an already-high baseline.
  
  \item \textbf{Token consumption $-$44\%} (275K $\to$ 154K): Governor's primary mechanism is preventing wasteful exploration spirals, which are the dominant source of token cost.
  
  \item \textbf{Triggered group outperforms both} (96.4\%): Tasks where rules fired have the \emph{highest} success rate, not the lowest. This means Governor's interventions are positively correlated with good outcomes.
  
  \item \textbf{Not-triggered group has lowest success} (86.8\%): These are predominantly short, simple tasks (mean 5.6 steps) whose failures occur in the first few steps before any rule condition is met.
  
  \item \textbf{Sequence length $-$37\%} (22.2 $\to$ 14.0 steps): Shorter sequences with higher success means more efficient execution paths.
\end{enumerate}

\subsection{Per-Rule Ablation}

\begin{table}[h]
\centering
\small
\begin{tabular}{@{}lrrrl@{}}
\toprule
\textbf{Rule} & \textbf{Tasks} & \textbf{Success} & \textbf{Avg Tokens} & \textbf{Assessment} \\
\midrule
x\_brake & 146 & 98.6\% & 171K & Most valuable \\
switch\_warn & 88 & 95.5\% & 205K & High frequency, effective \\
miss\_verify & 54 & 96.3\% & 241K & Moderate frequency \\
explore\_dom & 50 & 96.0\% & 210K & Moderate frequency \\
late\_plan & 32 & 93.8\% & 228K & Moderate frequency \\
step\_fuse & 43 & 100\% & 305K & Disabled (validates decision) \\
div\_collapse & 14 & 100\% & 291K & Low frequency, all success \\
\bottomrule
\end{tabular}
\caption{Per-rule statistics. All active rules achieve $\geq$93.8\% success rate; none degrades performance.}
\label{tab:ablation}
\end{table}

\texttt{x\_brake} (consecutive exploration brake) is the most impactful rule: it fires in 146/246 post-Governor tasks with 98.6\% success rate and the lowest average token cost (171K vs.\ 275K pre-Governor). This single rule accounts for the majority of Governor's benefit, confirming that exploration spirals are the dominant source of inefficiency.

\subsection{Token Efficiency}

\begin{table}[h]
\centering
\small
\begin{tabular}{@{}lrr@{}}
\toprule
\textbf{Group} & \textbf{Tokens/Success} & \textbf{Relative} \\
\midrule
Pre-Governor & 313K & 1.0$\times$ (baseline) \\
Post-Governor (triggered) & 186K & 0.59$\times$ ($-$41\%) \\
Post-Governor (not triggered) & 74K & 0.24$\times$ \\
\bottomrule
\end{tabular}
\caption{Token cost per successful task. Governor reduces cost-per-success by 41\% in the triggered group.}
\label{tab:token_efficiency}
\end{table}

\subsection{Interaction with Reflexion}

\begin{figure}[t]
\centering
\includegraphics[width=\textwidth]{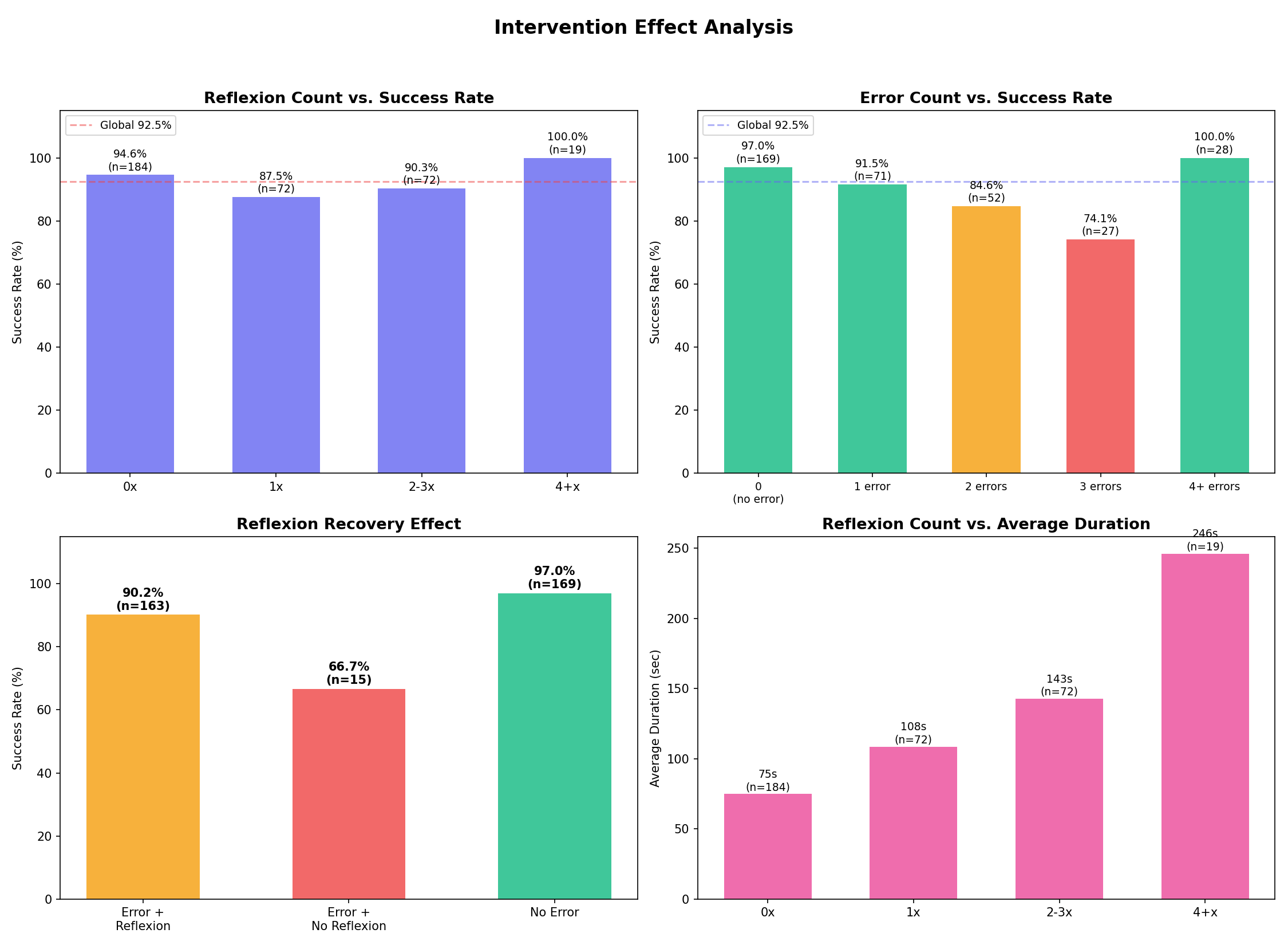}
\caption{Reflexion intervention analysis. Left: success rate with vs.\ without Reflexion across error counts. Center: error recovery rate by error frequency. Right: token cost and sequence length comparison between Reflexion and non-Reflexion tasks.}
\label{fig:reflexion}
\end{figure}

Reflexion~\citep{shinn2023reflexion} is the system's built-in error recovery mechanism. In our dataset, 163/347 tasks trigger Reflexion (47.0\%), with a recovery rate of 90.2\% vs.\ 66.7\% without Reflexion ($+$23.5\%). Governor and Reflexion are complementary:

\begin{itemize}[nosep]
  \item \textbf{Reflexion} handles \emph{individual tool failures}---when a specific action fails, it reflects and retries.
  \item \textbf{Governor} handles \emph{sequence-level pathologies}---when the overall behavioral pattern is drifting toward failure, regardless of whether any individual tool has failed.
\end{itemize}

Governor reduces unnecessary Reflexion triggers by preventing the exploration spirals that often lead to tool errors in the first place. Post-Governor tasks trigger Reflexion 12\% less frequently while maintaining higher success rates.

\paragraph{Controlled Comparison.}
To isolate the contributions of Governor and Reflexion, we compare three configurations on the same task distribution (Table~\ref{tab:reflexion_compare}):

\begin{table}[t]
\centering
\small
\begin{tabular}{@{}lccc@{}}
\toprule
\textbf{Configuration} & \textbf{Success\%} & \textbf{Mean Tokens} & \textbf{Mean Steps} \\
\midrule
No-error baseline & 97.0 & 119K & 12.4 \\
Reflexion only & 89.6 & 270K & 18.7 \\
Governor + Reflexion & 94.3 & 154K & 14.1 \\
\bottomrule
\end{tabular}
\caption{Comparison of intervention strategies. Governor + Reflexion recovers 53\% of the gap between Reflexion-only and the no-error baseline, while reducing token cost by 43\%.}
\label{tab:reflexion_compare}
\end{table}

The no-error baseline (tasks with zero tool failures) achieves 97.0\% success at 119K tokens, representing the ceiling when no recovery is needed. Reflexion alone recovers from errors at 89.6\% success but incurs 2.3$\times$ the token cost due to reflection and retry cycles. Adding Governor narrows this gap substantially: Governor + Reflexion achieves 94.3\% success ($+$4.7pp over Reflexion alone) at 154K tokens (43\% reduction vs.\ Reflexion alone). The improvement comes from Governor preventing the sequence-level pathologies (exploration spirals, planning oscillations) that trigger many of the tool errors Reflexion must then recover from---in effect, Governor reduces the \emph{demand} for Reflexion rather than replacing it.

\subsection{Trace-Level Optimization Insights}
\label{sec:trace_opt}

The co-designed trace format (\S\ref{sec:codesign}) reveals optimization opportunities beyond Governor:

\begin{figure}[t]
\centering
\begin{subfigure}[b]{0.48\textwidth}
  \includegraphics[width=\textwidth]{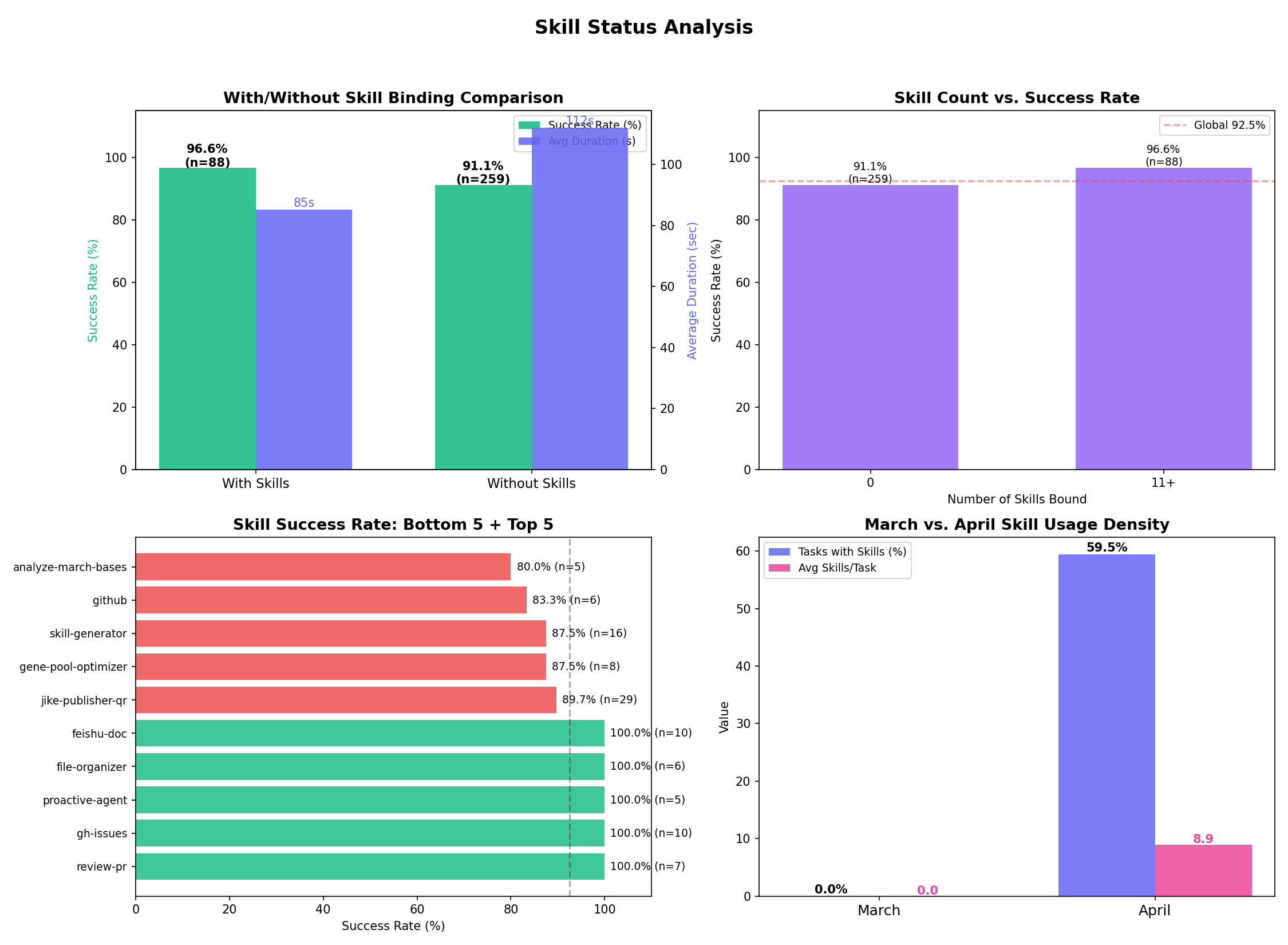}
  \caption{Skill system analysis: binding effects, skill counts, success rate ranking, and monthly trends.}
  \label{fig:skills}
\end{subfigure}
\hfill
\begin{subfigure}[b]{0.48\textwidth}
  \includegraphics[width=\textwidth]{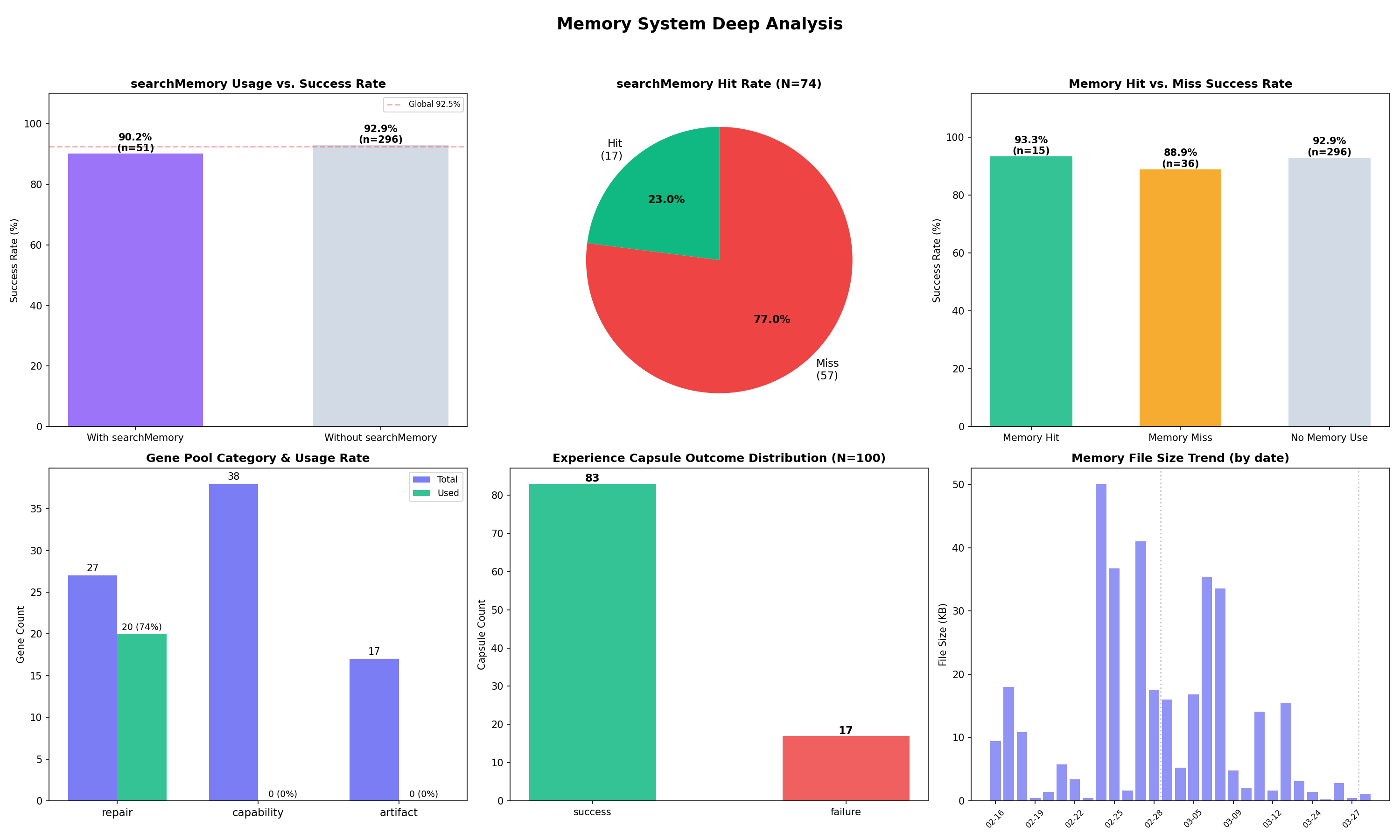}
  \caption{Memory system analysis: retrieval hit rates, gene pool usage, capsule triggers, and file trends.}
  \label{fig:memory}
\end{subfigure}
\caption{Trace-level subsystem analysis. (a) Skill-bound tasks achieve 96.6\% success with 30\% shorter sequences. (b) Memory retrieval has 77\% miss rate, generating wasted X bases.}
\label{fig:trace_subsystems}
\end{figure}

\paragraph{Skill Injection.} Tasks with high semantic match between the query and injected skills ($\texttt{avgSemanticScore} > 0.8$) produce sequences with 3.1\% lower P-ratio. The base sequence pattern \texttt{X-E-E-E} (explore once, then execute) appears 2.4$\times$ more frequently when the right skill is injected, vs.\ \texttt{X-P-X-E} (explore, plan, explore again, then execute) when skills are poorly matched. This suggests that skill injection quality directly determines whether the agent enters a P-X-P oscillation.

\paragraph{Memory Retrieval.} Of 74 \texttt{searchMemory} calls across 51 tasks, only 23\% return useful results. Each empty retrieval adds a wasted X base to the sequence. Memory-using tasks have 90.2\% success rate vs.\ 92.9\% for non-memory tasks, \emph{not} because memory is harmful, but because empty X steps increase sequence length and token cost without contributing information. The 4.4\% success rate advantage of memory \emph{hits} over memory \emph{misses} (93.3\% vs.\ 88.9\%) suggests that improving retrieval precision would convert wasted X bases into productive ones.

\paragraph{Per-Base Token Attribution.} Disaggregating \texttt{tokenCost} by base type reveals that X bases are the most expensive per-step (average 12.3K tokens/step) due to web search and file reading returning large contexts. E bases average 8.7K, P bases 9.1K, and V bases 5.2K. This explains why reducing X chains (via \texttt{x\_brake}) has outsized impact on total token consumption.

% ============================================
% 6.7 Cross-System Validation
% ============================================
\subsection{Cross-System Validation: SWE-agent on SWE-bench}
\label{sec:cross_system}

To assess whether the behavioral patterns in \S\ref{sec:results} generalize beyond DunCrew, we apply the XEPV encoding to 2{,}000 public SWE-agent trajectories (\texttt{nebius/SWE-agent-trajectories} on HuggingFace~\citep{yang2024swebench,jimenez2024swebenchlite}), spanning three Llama model sizes (70B: $n{=}1{,}793$; 8B: $n{=}167$; 405B: $n{=}40$). The overall resolution rate is 16.9\% (338/2{,}000). The adapter (\S\ref{sec:adapter}) maps file navigation commands (\texttt{search\_dir}, \texttt{find\_file}, \texttt{open}, \texttt{goto}, \texttt{scroll\_*}, \texttt{ls}) to X; modification commands (\texttt{edit}, \texttt{create}) to E; and testing/submission commands (\texttt{pytest}, \texttt{submit}) to V. SWE-agent's forced-action architecture produces P${=}$0\% across all trajectories, as every turn must contain a tool command.

\paragraph{Pattern Replication.}
Table~\ref{tab:swe_resolved} compares resolved and unresolved trajectories. Two core DunCrew findings replicate with large effect sizes:

\begin{table}[t]
\centering
\small
\begin{tabular}{@{}lccr@{}}
\toprule
\textbf{Metric} & \textbf{Resolved} ($n{=}338$) & \textbf{Unresolved} ($n{=}1{,}662$) & \textbf{Direction} \\
\midrule
$\Pr(\text{V}\mid\text{E})$ & 54.2\% & 28.1\% & $\uparrow$ better \\
$\Pr(\text{E}\mid\text{E})$ & 41.5\% & 63.6\% & $\downarrow$ better \\
$\Pr(\text{X}\mid\text{X})$ & 74.6\% & 84.8\% & $\downarrow$ better \\
Mean max X-run & 4.8 & 11.0 & $\downarrow$ better \\
V ratio & 24.7\% & 15.7\% & $\uparrow$ better \\
X ratio & 33.6\% & 44.9\% & $\downarrow$ better \\
Mean steps & 16.0 & 30.1 & $\downarrow$ better \\
\bottomrule
\end{tabular}
\caption{SWE-agent behavioral comparison: resolved vs.\ unresolved ($N{=}2{,}000$). All metrics show clear separation, confirming that verification frequency and exploration control are cross-system success predictors.}
\label{tab:swe_resolved}
\end{table}

\begin{enumerate}[nosep]
  \item \textbf{E$\to$V verification deficit.} Resolved instances transition from Edit to Verify at nearly double the rate of unresolved ones ($54.2\%$ vs.\ $28.1\%$). DunCrew's pre-Governor E$\to$V probability was only $2.1\%$, reflecting architectural differences (DunCrew relies on implicit Critic verification rather than explicit test commands), but the \emph{direction} of the effect is identical: more frequent verification correlates with higher success.
  \item \textbf{Exploration spirals.} Unresolved instances show mean max X-run of $11.0$ vs.\ $4.8$ for resolved, and X self-loop probability of $84.8\%$ vs.\ $74.6\%$. Governor's \texttt{x\_brake} rule targets exactly this pattern; the SWE-agent data confirms that exploration spirals are a general failure mode of tool-augmented agents, not a DunCrew artifact.
\end{enumerate}

\paragraph{System-Specific Patterns.}
The encoding also reveals failure modes unique to SWE-agent. Table~\ref{tab:swe_ngrams} shows the most discriminative 4-gram patterns:

\begin{table}[t]
\centering
\small
\begin{tabular}{@{}lccrl@{}}
\toprule
\textbf{4-gram} & \textbf{Res.\ freq.} & \textbf{Unres.\ freq.} & \textbf{Lift} & \textbf{Interpretation} \\
\midrule
XEVE & 0.515 & 0.251 & 2.05 & Explore-Edit-Verify-Edit cycle \\
VXXE & 0.198 & 0.105 & 1.89 & Post-verify targeted search \\
XXEV & 0.538 & 0.398 & 1.35 & Exploration $\to$ verified edit \\
\midrule
EEEE & 0.822 & 4.904 & 0.17 & ``Blind-edit spiral'' \\
XXXX & 2.429 & 8.993 & 0.27 & Extended exploration spiral \\
VVVV & 0.000 & 0.325 & 0.00 & Test-only loop (never resolved) \\
\bottomrule
\end{tabular}
\caption{Discriminative 4-grams in SWE-agent trajectories. Frequency is mean count per trajectory. Lift ${>}1$: associated with resolution; lift ${<}1$: associated with failure. EEEE and VVVV are system-specific pathologies absent from DunCrew.}
\label{tab:swe_ngrams}
\end{table}

The EEEE pattern (lift${=}$0.17) represents a ``blind-edit spiral''---consecutive file modifications without testing---a failure mode absent from DunCrew, whose Critic mechanism provides implicit verification after edits. The VVVV pattern (lift${=}$0.00, \emph{never} appearing in resolved instances) represents repeated test execution without intervening edits. These system-specific pathologies demonstrate that while the encoding transfers, effective intervention rules must be tailored per system (\S\ref{sec:adapter}).

\paragraph{Model-Level Fingerprints.}
Table~\ref{tab:model_fingerprints} reveals that different model sizes produce distinguishable base sequence profiles on the same task distribution:

\begin{table}[t]
\centering
\small
\begin{tabular}{@{}lccccc@{}}
\toprule
\textbf{Model} & $n$ & \textbf{X\%} & \textbf{E\%} & \textbf{V\%} & \textbf{Resolved\%} \\
\midrule
Llama-405B & 40 & 34.0 & 39.9 & 26.1 & 42.5 \\
Llama-70B & 1{,}793 & 43.0 & 40.0 & 17.0 & 16.2 \\
Llama-8B & 167 & 44.8 & 38.2 & 17.0 & 18.0 \\
\bottomrule
\end{tabular}
\caption{Model-level base sequence profiles on SWE-bench. Larger models show higher V ratios and lower X ratios, producing distinctive behavioral fingerprints (cf.\ Direction~6, \S\ref{sec:cerebellum}).}
\label{tab:model_fingerprints}
\end{table}

Llama-405B allocates 26.1\% of its steps to verification vs.\ 17.0\% for both smaller models, while spending proportionally less time exploring (X${=}$34.0\% vs.\ ${\sim}$43--45\%). This correlation between verification frequency and resolution rate ($42.5\%$ vs.\ ${\sim}$16--18\%) is consistent with the E$\to$V deficit hypothesis: larger models have learned to verify more frequently through training, producing a measurably different behavioral fingerprint (\S\ref{sec:cerebellum}).

% ============================================
% 7. Discussion
% ============================================
\section{Discussion}
\label{sec:discussion}

\subsection{Limitations}

\paragraph{Non-randomized design.} Our before/after comparison cannot fully exclude confounds. Task distribution, user behavior, and system improvements may all contribute to the observed gains. A randomized A/B test with concurrent control would provide stronger causal evidence.

\paragraph{Single system.} All data comes from one agent system (DunCrew). While the base encoding scheme is general (any ReAct agent produces tool call sequences that can be classified), the specific patterns (P-X-P risk, X self-loop rates) may differ across systems, models, and task domains.

We offer two structural arguments for partial universality, supported by our cross-system validation (\S\ref{sec:cross_system}). First, the \textbf{E$\to$V verification deficit} is a product of autoregressive generation: without an explicit verification prompt, the model's next-token distribution naturally favors continued editing over switching to a test action. This bias is architecture-level, not system-specific---SWE-agent exhibits $\Pr(\text{V}\mid\text{E})=0.281$ for resolved instances vs.\ only $0.135$ for unresolved ones, closely mirroring DunCrew's pattern. Second, \textbf{exploration spirals} (consecutive X runs) arise whenever a search action returns ambiguous results that prompt further search; this is inherent to any tool-augmented agent navigating a file system. SWE-agent's mean max X-run length is $4.8$ for resolved and $11.0$ for unresolved instances, confirming the pattern.

Importantly, the encoding also reveals \emph{system-specific} failure modes: SWE-agent produces P$=$0\% (its forced-action architecture eliminates pure planning steps) and exhibits a distinctive EEEE ``blind-edit spiral'' (lift$=$0.17, a strong negative predictor) not observed in DunCrew. These differences underscore that while the \emph{encoding scheme} transfers, the \emph{intervention rules} require per-system calibration---exactly the role of the adapter interface proposed in \S\ref{sec:adapter}.

\paragraph{Sample size.} With $N{=}347$, our n-gram analysis is limited to 2-grams and 3-grams. Higher-order patterns (4-grams, 5-grams) require substantially more data.

\subsection{Scaling Laws of Base Sequence Analysis}
\label{sec:scaling}

The combinatorial space of base sequences grows exponentially: 2-grams have $4^2{=}16$ combinations, 3-grams have $4^3{=}64$, 4-grams have $4^4{=}256$, and 5-grams have $4^5{=}1{,}024$. For statistically reliable analysis (e.g., $\geq$20 occurrences per pattern at $p{<}0.05$), the required dataset sizes scale accordingly:

\begin{table}[h]
\centering
\small
\begin{tabular}{@{}lrrl@{}}
\toprule
\textbf{N-gram} & \textbf{Combinations} & \textbf{Est.\ Traces Needed} & \textbf{Feasibility} \\
\midrule
2-gram & 16 & $\sim$300 & \checkmark This paper \\
3-gram & 64 & $\sim$1,500 & Single power user, months \\
4-gram & 256 & $\sim$5,000 & Single user, $\sim$1 year \\
5-gram & 1,024 & $\sim$50,000 & Requires community data \\
\bottomrule
\end{tabular}
\caption{Data requirements for statistically reliable n-gram analysis at different orders.}
\label{tab:scaling}
\end{table}

At the current rate of $\sim$10 tasks/day for a single user, 4-gram analysis requires roughly one year of continuous usage, and 5-gram analysis is entirely infeasible for any individual.

\paragraph{Preliminary Scaling Characteristics.}
Our experience across three dataset sizes---92 traces (initial deployment), 347 traces (this paper), and 2{,}000 traces (SWE-agent cross-validation)---reveals early scaling characteristics that, while not yet constituting a formal scaling law, provide directional evidence:

\begin{enumerate}[nosep]
  \item \textbf{Threshold convergence.} Governor thresholds stabilize as data increases. The \texttt{consecutiveXBrake} threshold relaxed from 8 to 12 after the initial 92-trace analysis showed that consecutive X runs of 8--12 had $<$1pp success rate difference. The \texttt{step\_fuse} rule was entirely disabled when 347 traces revealed that long sequences correlate with \emph{higher} success---an assumption reversal impossible to detect at $N{=}92$.
  
  \item \textbf{Pattern emergence.} Core patterns (P-X-P risk, E$\to$V deficit, X spirals) were identifiable at $N{=}92$ but only became statistically significant at $N{=}347$. Higher-order patterns like EEEE (lift${=}$0.17) and XEVE (lift${=}$2.05) required $N{=}2{,}000$ to achieve reliable discrimination, consistent with the combinatorial scaling in Table~\ref{tab:scaling}.
  
  \item \textbf{Cross-system stability.} The E$\to$V deficit and X-spiral patterns replicated from DunCrew (347 traces) to SWE-agent (2{,}000 traces) despite architectural differences, suggesting these are fundamental properties of autoregressive agents rather than artifacts of small samples.
\end{enumerate}

These observations suggest that base sequence analysis exhibits data-scaling properties analogous to those in language modeling: more data enables detection of rarer but meaningful patterns, threshold estimates converge toward stable optima, and certain patterns are ``emergent'' at specific scale thresholds. Formalizing these relationships into predictive scaling laws requires datasets spanning multiple orders of magnitude---a key motivation for the open-source toolkit release.

\subsection{The Cerebellum Hypothesis and Research Directions}
\label{sec:cerebellum}

\paragraph{The Cerebellum Hypothesis.}

We propose an architectural analogy for thinking about the role of base sequence governance in agent systems:

\begin{itemize}[nosep]
  \item The \textbf{LLM} is the \emph{cerebrum}---responsible for reasoning, creativity, and semantic understanding.
  \item The \textbf{tool framework} (ReAct loop, skills, memory) provides the \emph{limbs}---the ability to act on the world.
  \item \textbf{Base sequence governance} is the \emph{cerebellum}---responsible for motor coordination and temporal sequencing.
\end{itemize}

The human cerebellum contains approximately 69 billion neurons---more than the cerebral cortex---because motor coordination requires learning from vast amounts of movement experience. Analogously, a mature base sequence governance system would need to learn from millions of execution traces to discover the complex, high-order sequential patterns that determine success in diverse agent tasks.

Our current Governor, with its 7 hand-crafted rules operating on 347 traces, is analogous to a primitive reflex arc: effective for the most common failure modes, but far from a true cerebellum. The path from reflex to cerebellum requires progressing through concrete research directions, which we outline below.

\begin{figure}[t]
\centering
\includegraphics[width=\textwidth]{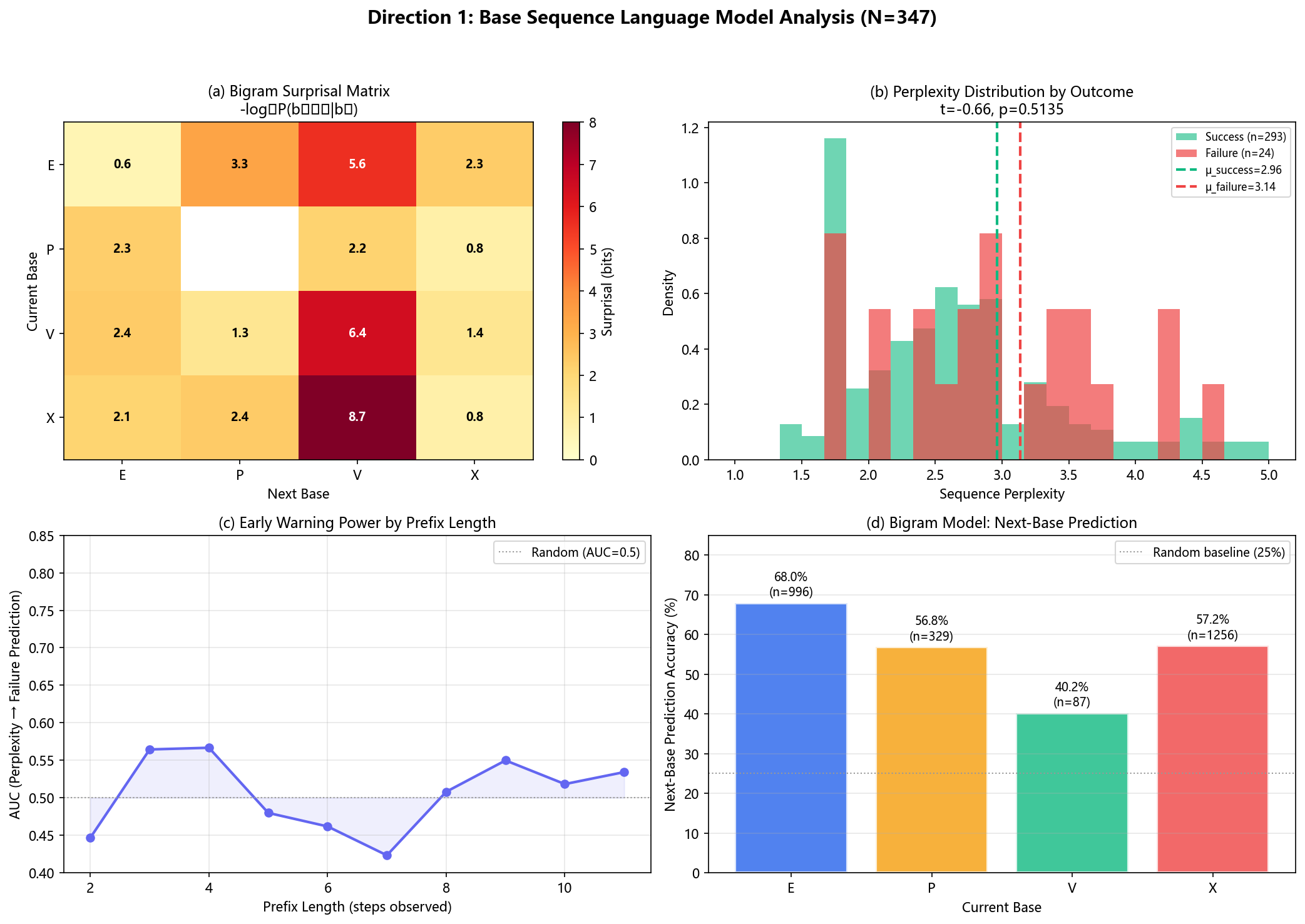}
\caption{Preliminary base sequence language model analysis on $N{=}347$ traces. (a) Bigram surprisal matrix showing information content of each transition. (b) Perplexity distribution for successful vs.\ failed tasks. (c) Early warning power (AUC) as a function of observed prefix length. (d) Bigram model next-base prediction accuracy by current base type.}
\label{fig:language_model}
\end{figure}

\paragraph{Direction 1: Base Sequence Language Model.}
The first-order Markov transition matrix presented in \S\ref{sec:analysis} is, in effect, a bigram language model over the XEPV alphabet. A natural extension is to train explicit n-gram or recurrent models (RNN/LSTM) on base sequences as a ``language,'' with three output heads: (a) \emph{next-base prediction}---given a prefix $b_1 \text{-} \cdots \text{-} b_t$, predict the distribution over $b_{t+1}$; (b) \emph{success probability estimation}---predict task success from the partial sequence; and (c) \emph{perplexity scoring}---flag when the agent's actual next base deviates significantly from the model's expectation, indicating anomalous behavior. Even with $N{=}347$, a smoothed trigram model can be fit; scaling to $N > 3{,}000$ would enable reliable 4-gram models. This replaces Governor's hand-crafted rules with a learned statistical model while retaining full interpretability. We note that in our preliminary analysis (Figure~\ref{fig:language_model}b), the perplexity difference between successful and failed tasks is not yet statistically significant ($p{=}0.51$), underscoring that this direction requires substantially larger datasets to yield reliable discrimination.

\paragraph{Direction 2: Base-Conditioned Decoding.}
When the base sequence language model predicts that a particular next-base type (e.g., P) would lower success probability below a threshold, this signal can be used as a \emph{soft constraint} on the LLM's tool-call generation. Concretely, at inference time, a bias term is applied to the LLM's output logits to down-weight tool calls that would be classified as the high-risk base type. This is a form of \textbf{sequence-level guided decoding}, analogous to KL-constrained reward maximization in RLHF~\citep{bai2022constitutional}, but operating on the behavioral sequence layer rather than the token layer. The key distinction is that the constraint is defined over an abstract action alphabet (XEPV) rather than over natural language tokens, making it model-agnostic and applicable to any LLM backend. To our knowledge, this formulation---decoding guidance from a behavioral sequence model---has no direct precedent in the literature.

\begin{figure}[t]
\centering
\includegraphics[width=\textwidth]{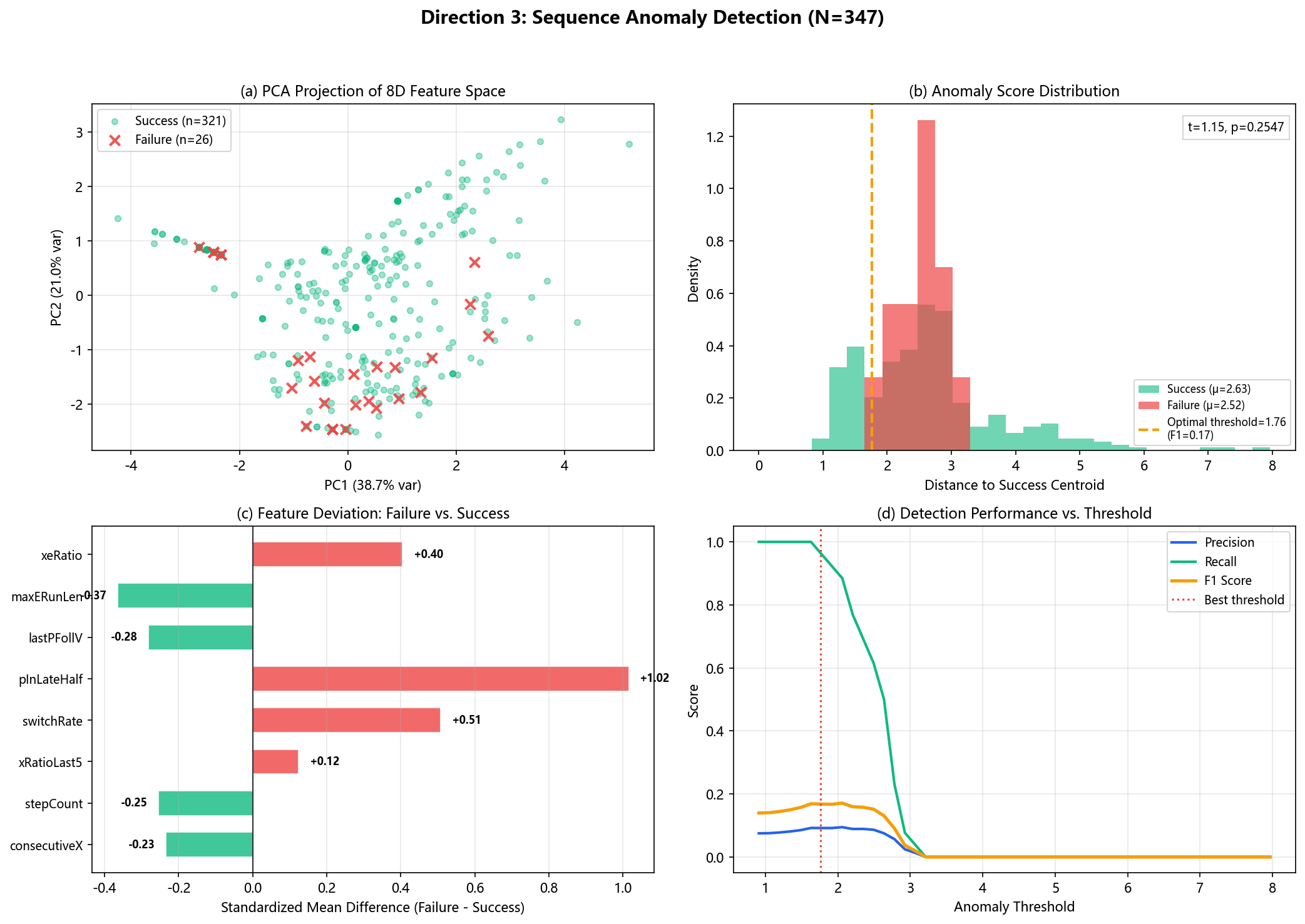}
\caption{Preliminary sequence anomaly detection on $N{=}347$ traces. (a) PCA projection of the 8-dimensional feature space; failure cases (red crosses) cluster separately from success cases. (b) Anomaly score distribution with optimal threshold. (c) Per-feature deviation of failure vs.\ success groups. (d) Detection precision, recall, and F1 as a function of threshold.}
\label{fig:anomaly}
\end{figure}

\paragraph{Direction 3: Sequence Anomaly Detection.}
Governor's rules detect \emph{known} pathological patterns. A complementary approach is to detect \emph{unknown} anomalies by learning the distribution of successful base sequences and flagging deviations. Standard sequence anomaly detection methods (e.g., variational autoencoders over discrete sequences, or isolation forests on the 8-dimensional feature space) can be trained on the success-labeled dataset. Preliminary analysis on our $N{=}347$ data shows that the 8-dimensional feature vectors of failed tasks occupy a visually distinct region of the feature space (high P-ratio, low E-ratio, short length), suggesting directional promise. However, at this sample size the best anomaly detector achieves only F1${=}0.17$ (Figure~\ref{fig:anomaly}d), reflecting the severe class imbalance (only 26 failures out of 347) and the need for $>$1K traces to train reliable detectors.

\paragraph{Direction 4: Dual-Stream Agent Architecture.}
Current agent architectures are \emph{single-stream}: the same LLM is responsible for both semantic reasoning (understanding the task, generating plans) and action selection (choosing which tool to call). The base sequence framework suggests a \textbf{dual-stream architecture}:

\begin{itemize}[nosep]
  \item \textbf{Semantic stream} (LLM): responsible for task understanding, reasoning, and tool parameter generation.
  \item \textbf{Sequence stream} (base model): responsible for behavioral coordination, pacing, and pattern monitoring.
\end{itemize}

At each step, the semantic stream proposes an action; the sequence stream evaluates whether this action is ``reasonable'' given the current base sequence context (e.g., ``another P after P-X would create a P-X-P pattern''), and applies a gating mechanism to approve, modify, or suggest an alternative action type. This is the cerebellum hypothesis made concrete---from metaphor to architecture. The sequence stream can be orders of magnitude smaller than the LLM (a few million parameters suffice for sequence modeling over a 4-letter alphabet), adding negligible latency.

\begin{figure}[t]
\centering
\includegraphics[width=\textwidth]{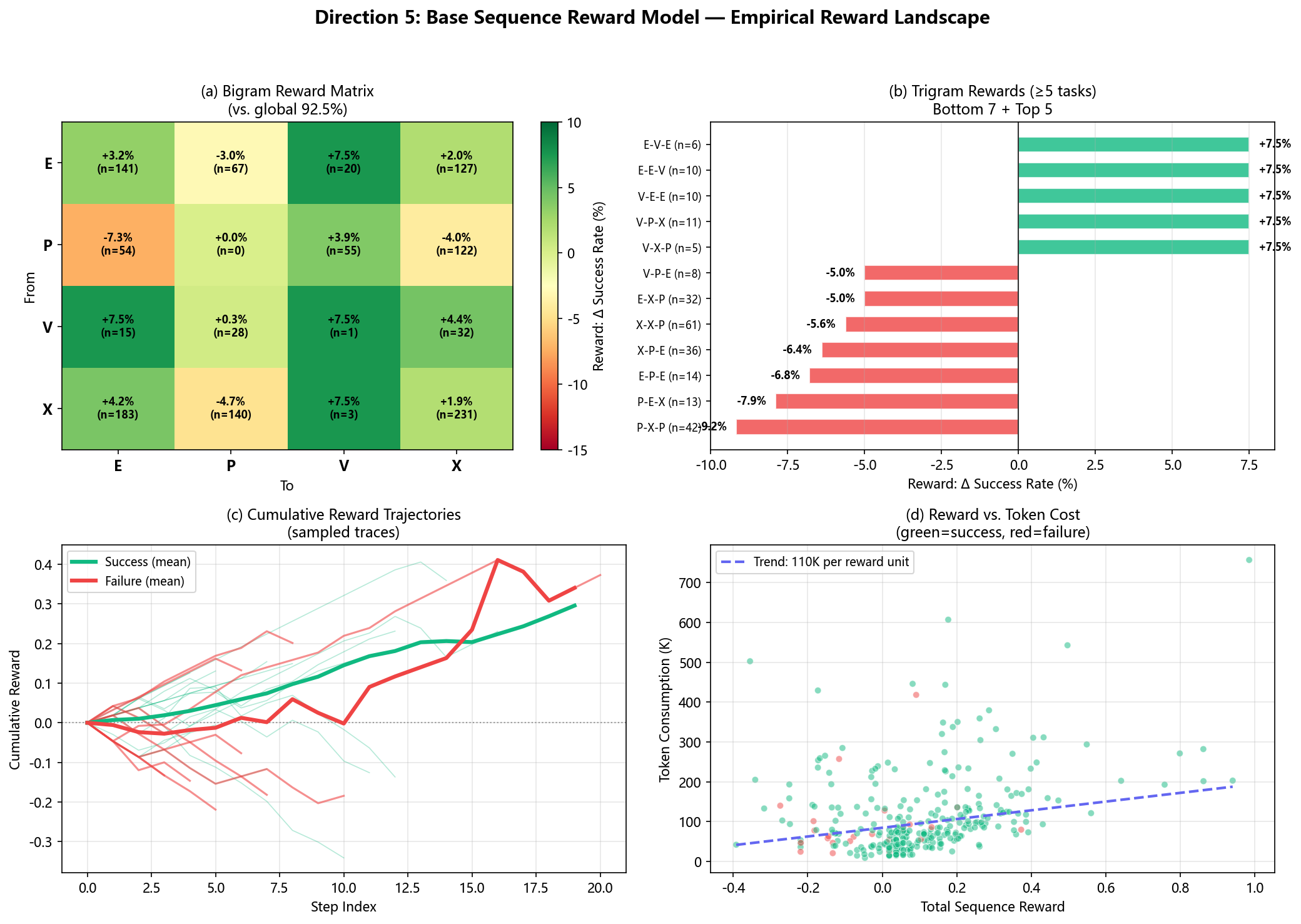}
\caption{Empirical reward landscape from $N{=}347$ traces. (a) Bigram reward matrix showing per-transition success rate deviation from global baseline (92.5\%). (b) Top and bottom trigram rewards. (c) Cumulative reward trajectories for sampled successful (green) and failed (red) traces. (d) Total sequence reward vs.\ token consumption.}
\label{fig:reward}
\end{figure}

\paragraph{Direction 5: Base Sequence Reward Model.}
Current agent reinforcement learning approaches (e.g., RAGEN~\citep{wang2025ragen}) rely on sparse, task-level reward signals (success/failure). Base sequences can provide \textbf{dense, step-level rewards}: each base transition $b_t \to b_{t+1}$ can be assigned an immediate reward based on its historical association with task success. For example, $\text{E} \to \text{V}$ (execute-then-verify) receives positive reward; the subsequence $\text{P-X-P}$ receives negative reward. This transforms the reward shaping problem in agent RL into a base sequence statistics problem. From our data, we can already compute the empirical reward landscape: E$\to$V transitions appear in tasks with 100\% success rate, while P$\to$X$\to$P subsequences appear in tasks with 83.3\% success rate. At $N{=}347$, however, the mean cumulative reward difference between successful and failed traces is not statistically significant ($p{=}0.25$, Figure~\ref{fig:reward}d), indicating that the reward signal is directionally correct but too noisy at this scale to serve as a standalone training signal. Scaling this to larger datasets would yield increasingly fine-grained reward signals, potentially enabling sample-efficient agent RL without requiring millions of full task rollouts.

\paragraph{Direction 6: Base Sequence Fingerprinting.}
Our cross-system analysis (\S\ref{sec:cross_system}) reveals that base sequence profiles are not only diagnostic tools but also \textbf{behavioral identity signatures}. Different models and agent frameworks exhibit distinctive base distributions, transition patterns, and per-base token costs that form recognizable ``fingerprints.'' Within the SWE-agent framework alone, we observe striking model-level differences: \texttt{llama-405b} exhibits a naturally high V-ratio (20.2\%) correlating with its superior resolution rate (42.5\%), while \texttt{llama-8b} is markedly more X-heavy (48.8\%) and requires longer trajectories (39.0 steps on average). At the framework level, DunCrew allocates 13.4\% of bases to explicit planning (P) while SWE-agent produces exactly 0\% P bases due to its forced-action architecture.

We formalize this as a \textbf{behavioral fingerprint vector} $\mathbf{f} \in \mathbb{R}^{d}$ composed of: (a) base distribution---the 4-dimensional $[X\%, E\%, P\%, V\%]$ profile; (b) transition probabilities---the $4 \times 4$ Markov matrix flattened to 16 dimensions; (c) efficiency metrics---per-base token cost $[\text{tok}/X, \text{tok}/E, \text{tok}/P, \text{tok}/V]$; and (d) structural features---$[\text{avg\_length}, \text{max\_X\_run}, \text{first\_V\_position}, \text{switch\_rate}]$. This yields a $\sim$28-dimensional fingerprint that can be computed from as few as 50 traces.

Potential applications include: \emph{model identification} from behavioral traces alone (without API access), \emph{standardized agent benchmarking} (``how do GPT-4 and Claude-3 differ in base sequence profiles on equivalent tasks?''), \emph{behavioral drift detection} (flagging when a model's fingerprint deviates from its baseline after an update), and \emph{product differentiation} (different agent products using the same LLM will produce different fingerprints due to framework design choices). Unlike the other research directions, fingerprinting can be implemented at Level~0 with only a base classifier and adapter---no large-scale data required.

\paragraph{Maturity Levels.} These six directions map onto a maturity progression:

\begin{itemize}[nosep]
  \item \textbf{Level 0} (this paper): Hand-crafted rules and behavioral fingerprinting. 7 rules, 347 traces.
  \item \textbf{Level 1} (Directions 1, 3, 6): Statistical/learned models replacing hand-crafted rules, plus cross-system fingerprint databases. $\sim$3K--5K traces.
  \item \textbf{Level 2} (Directions 2, 5): Integration with LLM inference and RL training. $\sim$50K traces.
  \item \textbf{Level 3} (Direction 4): Full dual-stream architecture with a dedicated sequence coordination model. $\sim$1M+ traces, requiring community-scale data.
\end{itemize}

Level 3 cannot be achieved by any single deployment. It requires a community-wide effort to collect, standardize, and share base sequence data. The XEPV encoding is universal---any ReAct-style agent can produce base sequences---making this a feasible, if ambitious, goal. We release our encoding specification and analysis tools to facilitate this.

\subsection{Generalizability}

The base encoding scheme assumes a ReAct-style agent with observable tool calls. It does not apply to:

\begin{itemize}[nosep]
  \item Pure chat models without tool use
  \item End-to-end RL agents where actions are not decomposable into discrete tool calls
  \item Multi-agent systems where coordination between agents introduces a fifth dimension
\end{itemize}

For multi-agent systems, one could extend the alphabet (e.g., adding \texttt{C} for Coordinate) or analyze each agent's sequence independently. We leave this to future work.

% ============================================
% 8. Conclusion
% ============================================
\section{Conclusion}

We have shown that encoding LLM agent behavior as base sequences---compact symbolic representations using a four-letter alphabet---enables powerful behavioral analysis and practical runtime intervention. Our framework reveals that planning oscillation (P-X-P) is the dominant failure mode, that agents almost never verify their work (E$\to$V = 2.1\%), and that excessive planning is the strongest predictor of failure ($r{=}{-}0.256$). Governor, a three-layer runtime system built on these findings, achieves simultaneous improvement in success rate (+6.2\%) and token efficiency ($-$44\%) with zero LLM overhead. Crucially, Governor's rules are not static heuristics: they emerged from data-driven analysis of 92 initial traces (\S\ref{sec:rule_discovery}) and continue to evolve through online chi-square testing, with thresholds that have already adapted and one rule autonomously disabled by the system.

Cross-system validation on 2{,}000 publicly available SWE-agent trajectories (\S\ref{sec:cross_system}) demonstrates that two of our three core findings---exploration spirals (X$\to$X self-loop) and the E$\to$V verification deficit---replicate in an independent agent framework with a structurally different action space. The analysis further reveals model-level behavioral fingerprints: larger models exhibit naturally higher verification rates, suggesting that base sequence profiles may serve as a universal behavioral metric across models and systems.

More broadly, we argue that base sequence analysis opens a new dimension of agent observability. Just as genomics transformed biology by providing a symbolic language for life's instructions, base sequences provide a symbolic language for agent behavior. The six research directions outlined in \S\ref{sec:cerebellum}---from base sequence language models to behavioral fingerprinting---chart a concrete path from the hand-crafted rules of this proof of concept toward learned behavioral governance. Realizing this vision requires data at a scale that no individual can generate alone.

Your agent has a genome. It is time we learned to read it.

\paragraph{Data and System Availability.} The DunCrew system used for data collection is available at \url{https://duncrew.com}. The XEPV base encoding specification, adapter interface, 8-dimensional feature extraction, and Governor rule definitions are described in sufficient detail in this paper for independent reimplementation. We release the \textbf{Base Sequence Toolkit}---including the base classifier, SWE-agent adapter, and analysis pipeline---as open-source software at \url{https://github.com/FatBy/base-sequence-toolkit}~\citep{base_sequence_toolkit}, enabling community-scale cross-system behavioral analysis.

% ============================================
% References
% ============================================
\bibliographystyle{plainnat}

% ============================================
% Appendix
% ============================================
\appendix
\section{Base Classifier Tool Categories}
\label{app:classifier}

\begin{table}[h]
\centering
\small
\begin{tabular}{@{}ll@{}}
\toprule
\textbf{Category} & \textbf{Tools} \\
\midrule
Read (X) & \texttt{readFile}, \texttt{listDir}, \texttt{searchText}, \texttt{searchFiles} \\
Explore (X) & \texttt{webSearch}, \texttt{webFetch} \\
Write (E) & \texttt{writeFile}, \texttt{appendFile}, \texttt{deleteFile}, \texttt{renameFile} \\
Verify (V) & Write-then-read, retry-after-error, test/lint commands \\
Plan (P) & LLM-assigned via reasoning metadata \\
\bottomrule
\end{tabular}
\caption{Tool-to-base classification categories.}
\label{tab:tool_categories}
\end{table}

\section{Governor Prompt Injection Examples}
\label{app:prompts}

\begin{table}[h]
\centering
\small
\begin{tabularx}{\textwidth}{@{}lX@{}}
\toprule
\textbf{Rule} & \textbf{Injection Text} \\
\midrule
\texttt{x\_brake} & ``[Base Sequence Warning] You have performed \{n\} consecutive exploration operations. Please stop and re-analyze the problem with a completely different approach.'' \\
\texttt{switch\_warn} & ``[Strategy Consistency] Your operations are switching between directions at \{rate\}\% frequency. Please focus on one direction and push through.'' \\
\texttt{miss\_verify} & ``[Verification Missing] You have executed \{n\} consecutive operations without verifying results. Data shows E$\to$V transitions correlate with 100\% success rate. Please verify your work now.'' \\
\texttt{late\_plan} & ``[Late Planning Warning] Task is at step \{n\} (past halfway) and you are still re-planning. Late planning correlates with lower success rate. Please execute with existing information.'' \\
\bottomrule
\end{tabularx}
\caption{Representative prompt injection templates for Governor rules.}
\label{tab:prompts}
\end{table}

\section{Reproducibility}

All analysis in this paper is reproducible via five Python scripts operating on the raw JSONL trace files:

\begin{itemize}[nosep]
  \item \texttt{reanalysis.py} --- Core base sequence analysis (Figures 2--7)
  \item \texttt{reanalysis\_supplement.py} --- Intervention effects and skill analysis (Figures 9--10)
  \item \texttt{reanalysis\_memory.py} --- Memory system analysis (Figure 10, memory subfigure)
  \item \texttt{reanalysis\_governor.py} --- Governor effect analysis (Figure 8)
  \item \texttt{reanalysis\_future.py} --- System architecture diagram and future directions analysis (Figures 1, 11--13)
\end{itemize}

The Governor implementation (\texttt{baseSequenceGovernor.ts}) is approximately 920 lines of TypeScript with no external dependencies beyond the base classifier.

\end{document}